\documentclass{article}

 \usepackage[preprint]{neurips_2026}

\usepackage[utf8]{inputenc} 
\usepackage[T1]{fontenc}    
\usepackage{hyperref}       
\usepackage{url}            
\usepackage{booktabs}       
\usepackage{colortbl}       
\usepackage{amsfonts}       
\usepackage{amsthm}         
\usepackage{nicefrac}       
\usepackage{siunitx}
\usepackage{colortbl}       
\usepackage{graphicx}       
\usepackage{subcaption}    
\usepackage{wrapfig}        
\usepackage{microtype}      
\usepackage{xcolor}         
\usepackage{amsmath}
\usepackage{amssymb}        
\usepackage{mathtools}
\usepackage{caption}
\usepackage{needspace}
\usepackage[linesnumbered,ruled,vlined]{algorithm2e}
\usepackage{algpseudocode}
\usepackage{multirow}
\usepackage{placeins}
 
\usepackage[table]{xcolor} 
\usepackage{array}
\definecolor{methodbg}{gray}{0.94}
\usepackage[capitalize,noabbrev]{cleveref}
\usepackage[dvipsnames]{xcolor}         
\usepackage{enumitem}
\definecolor{myblue}{RGB}{16, 72, 245}
\hypersetup{
    colorlinks=true,
    linkcolor=Maroon,
    citecolor=blue,      
    urlcolor=Blue, 
    }

\title{Noise Scheduling as \\Information-Guided Allocation in Diffusion Training}

 \author{%
Gabriel Raya$^{1}$\thanks{Work partially done during an internship at Sony AI, Tokyo, Japan.}\quad
Bac Nguyen$^{2}$\quad
Georgios Batzolis$^{3}$\quad
Yuhta Takida$^{2}$\quad
Dejan Stancevic$^{4}$\\
\textbf{Naoki Murata}$^{2}$\quad
\textbf{Chieh-Hsin Lai}$^{2}$\quad
\textbf{Yuki Mitsufuji}$^{2,5}$\thanks{These authors contributed equally as senior authors.}\quad
\textbf{Luca Ambrogioni}$^{4}$\footnotemark[2]\\
$^{1}$Tilburg University \& JADS\quad
$^{2}$Sony AI\quad
$^{3}$University of Cambridge\\
$^{4}$Radboud University\quad
$^{5}$Sony Group Corporation.
}

\definecolor{burntorange}{rgb}{0.80,0.33,0.00}
\definecolor{ForestGreen}{rgb}{0.13,0.55,0.13}
\definecolor{RoyalPurple}{rgb}{0.47,0.32,0.66}
\definecolor{Fuchsia}{HTML}{EE2967}

\definecolor{Wcol}{RGB}{219,234,254}
\definecolor{PHIcol}{RGB}{254,235,200}

\definecolor{hirow}{HTML}{FFF4D6}
\definecolor{mistrow}{HTML}{EAF2F8}
\definecolor{sagerow}{HTML}{EEF3EC}
\definecolor{paperrow}{HTML}{F4F4F4}

\definecolor{methodbg}{gray}{0.94}

\newcommand{\E}{\mathbb{E}}
\DeclareMathOperator{\Var}{Var}
\DeclareMathOperator{\Cov}{Cov}
\DeclareMathOperator{\tr}{tr}

\newcommand{\dd}{\mathrm{d}}

\newcommand{\defeq}{\triangleq}
\DeclareMathOperator{\mmse}{mmse}

\newcommand{\norm}[1]{\left\lVert #1 \right\rVert}

\newcommand{\I}{\mathbf{I}}

\newcommand{\btheta}{\boldsymbol{\theta}}

\newcommand{\rvx}{\mathbf{x}}

\newcommand{\rvz}{\mathbf{z}}

\newcommand{\bepsilon}{\boldsymbol{\varepsilon}}

\newcommand{\xhat}{\hat{\rvx}_{\btheta}}

\newcommand{\method}{\textsc{InfoNoise}}

\newcommand{\Hent}{\mathrm{H}}

\DeclareMathOperator{\sech}{sech}

\theoremstyle{plain}

\theoremstyle{definition}

\theoremstyle{remark}

\begin{document}

\maketitle

\begin{abstract}
We introduce \method{}, an online adaptive noise schedule for diffusion training that reallocates optimization effort toward noise levels where denoising is most informative. Together with loss weighting, a noise schedule induces an effective allocation across denoising problems, often fixed before informative noise levels are known. \method{} makes this allocation data-adaptive by estimating a conditional-entropy-rate profile from denoising losses during training, without auxiliary models or offline search. Through I--MMSE, this profile identifies where noisy observations rapidly reduce uncertainty about the clean sample and guides adaptation of the training noise distribution. It changes only this distribution, keeping the objective, weighting, and parameterization fixed. On image benchmarks, where schedules have been extensively tuned, \method{} matches or slightly exceeds strong baselines and can reach the same quality with fewer updates. On representation, sequence, and modality shifts, including DNA and language generation, \method{} improves over fixed and adaptive baselines and reaches target quality with up to \(3\times\) less training compute. These results establish the conditional-entropy-rate profile as the data-dependent target for noise schedule design and make online adaptation a practical alternative to manual schedule search.
\end{abstract}
\section{Introduction}

Diffusion models have become a dominant class of generative models, with strong performance across images, video, audio, language, and scientific data~\citep{nichol2021improved,rombach2022high,ho2022video,kong2020diffwave,he2023diffusionbert,kahouli2024molecular,lai2025principles}. They generate data by learning to reverse a corruption process that gradually obscures structure with noise~\citep{sohl2015deep,song2019generative,ho2020denoising,song2020score}. Training this reverse process does not optimize a single denoising problem, but a weighted collection of denoising losses across noise levels. Each noise level induces a different posterior inference problem, with different remaining uncertainty and a different learning signal. The training noise schedule sets how often each problem is visited, while loss weighting sets how strongly its error enters the objective. Together they induce an \emph{effective allocation} of training effort across the corruption path~\citep{karras2022elucidating,kingma2023understanding}. The central challenge is that this allocation is usually fixed before training reveals where along the path noisy observations are most informative.

In practice, diffusion training has often improved by changing this allocation indirectly. Noise schedules, loss weights, denoising objectives, and prediction parameterizations change which regions of the corruption path receive optimization effort~\citep{nichol2021improved,karras2022elucidating,hang2023efficient,kingma2023understanding,karras2024analyzing,esser2024scaling}. Yet the informative region can move with resolution, representation, data statistics, or modality, even within images~\citep{hoogeboom2023simple}; practice then relies on inherited schedules, empirical search, or objective-tied adaptive signals, while the allocation target remains implicit~\citep{austin2021structured,he2023diffusionbert,shi2024simplified,sahoo2024simple,chen2026langflow}.
When a schedule tuned in one regime is reused in another, it transfers the source allocation rather than measuring which noise levels are informative for the current training distribution. If the informative region shifts, inherited allocation remains tied to the source regime and spends updates where noisy observations resolve little uncertainty. We call this failure mode \emph{allocation mismatch}. This raises the question of whether training allocation can adapt online to the data, rather than be inherited or recovered through empirical search.

The allocation target can be made precise by asking where uncertainty is resolved along the corruption path. Conditional entropy measures the uncertainty about the clean sample that remains after a noisy observation~\citep{shannon1948mathematical}; its pathwise rate identifies where observations reduce that uncertainty most rapidly. For affine-Gaussian corruption paths, I--MMSE links this rate to Bayes-optimal denoising error~\citep{guo2005mutual,palomar2005gradient}. The resulting conditional-entropy-rate profile gives a data-dependent target for training allocation. Schedule tuning can be understood as an indirect search for this profile. Fixed schedules are effective when they place effort near it, and inefficient when the profile shifts while allocation remains inherited. Thus, the governing object for schedule design is not the schedule family itself, but the information profile over noise levels.

\begin{figure*}[t]
  \centering
  \includegraphics[width=\linewidth]{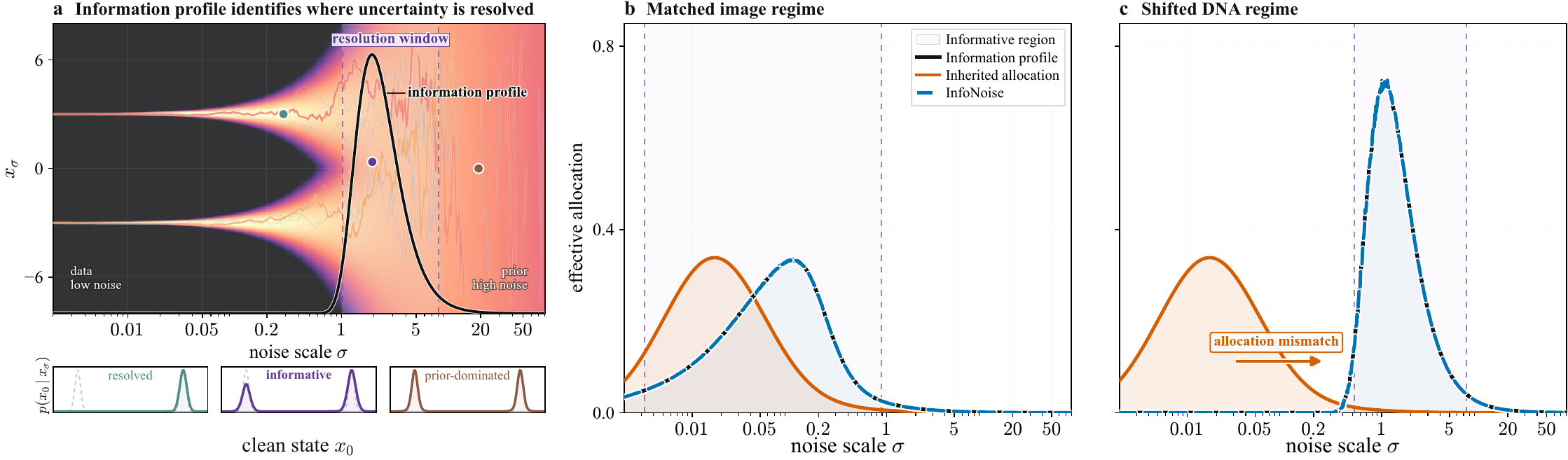}
\caption{
\textbf{The information profile defines the allocation target in diffusion training.}
\textbf{(a)} In an analytical two-state Gaussian channel, uncertainty is not resolved uniformly along the path. The entropy-rate profile peaks at the decision window where noisy observations most rapidly reveal the clean state; the derivation is given in \cref{app:toy_symmetry}.
\textbf{(b)} In mature image regimes, inherited allocations already overlap the measured informative region, so \method{} preserves the allocation while recovering this region online.
\textbf{(c)} Under shift, the information profile moves while inherited allocation remains fixed, producing allocation mismatch. \method{} estimates the shifted profile online and redirects allocation toward the region where uncertainty is resolved, without noise-schedule retuning.
}
  \label{fig:allocation_mismatch}
  \vspace{-1.8em}
\end{figure*}

We introduce \method{}, an online noise-schedule adaptation method that estimates this information profile from denoising losses during training and updates only the training noise distribution. The objective, loss weighting, and prediction parameterization remain fixed, so the comparison isolates training allocation from changes to the learning problem. Unlike entropic-time methods that reparameterize inference-time discretization of a pretrained model~\citep{stancevic2025entropic}, \method{} adapts the training allocation itself. \Cref{fig:allocation_mismatch} illustrates the allocation view. Along the corruption path, uncertainty about the clean sample is resolved non-uniformly, so efficient training should concentrate effort where denoising carries the most information. Panel~(a) shows a controlled two-state Gaussian channel in which the conditional-entropy-rate profile localizes the resolution window. Panel~(b) shows the matched case, where tuned image allocations overlap the informative region and \method{} recovers this region online. Panel~(c) shows the shifted case, where the informative region moves while inherited allocation remains fixed, producing allocation mismatch. \method{} uses the recovered information profile to adapt training to the denoising problem itself, rather than to an inherited schedule, moving updates toward the regions where uncertainty is resolved.

\textbf{Contributions.}
\textbf{(i)} We formalize diffusion noise scheduling as an effective allocation problem along the corruption path.
\textbf{(ii)} We identify the conditional-entropy-rate profile as the data-dependent target for this allocation, explaining when fixed schedules transfer and when they misallocate compute.
\textbf{(iii)} We introduce \method{}, an online adaptive noise schedule that estimates this profile from denoising losses and updates only the training noise distribution, replacing repeated schedule search with online information-guided allocation.
\section{Effective allocation and the information profile}
\label{sec:allocation_view}

Let \(p_{\mathrm{data}}\) denote the distribution of clean samples on \(\mathbb R^d\), and let \(\rvx_0\sim p_{\mathrm{data}}\). Diffusion models define a corruption path that turns clean samples into noisy observations and train a reverse model through the denoising problems induced along it. The path is often parameterized by \(t\in[0,1]\) and a corruption schedule, such as a noise scale \(\sigma(t)\), a monotone transform of log-SNR, or signal and noise coefficients \(a(t)\) and \(b(t)\). For allocation, the relevant object is the ordered family of denoising problems, not the coordinate used to label it. We write \(u\) for the coordinate in which allocation is analyzed, chosen in the data-to-noise direction. For the affine-Gaussian paths considered here,
\begin{equation}
    \rvx_u
    =
    a(u)\rvx_0 + b(u)\bepsilon,
    \qquad
    \bepsilon\sim\mathcal N(\mathbf 0,\I),
\label{eq:affine_gaussian_channel}
\end{equation}
where \(a(u)\) and \(b(u)\) set the signal and noise strengths. For all path values used below, \(a(u),b(u)>0\), and the signal-to-noise ratio \(\gamma(u)=a(u)^2/b(u)^2\) is non-increasing in \(u\). Each value of \(u\) defines the posterior denoising problem of estimating \(\rvx_0\) from \(\rvx_u\). In VE paths, \(u=\sigma\), \(a(u)\equiv1\), and \(b(u)=u\), giving \(\rvx_\sigma=\rvx_0+\sigma\bepsilon\). Standard VE, VP, and sub-VP corruption kernels fit this form, as do the Gaussian conditional interpolants used in our flow-matching baselines~\citep{song2020score,karras2022elucidating,kingma2023understanding,lipman2023flow}.

\subsection{Effective allocation}
\label{sec:training_allocation}

Let \(\pi(u)\) denote the training density over path locations in the coordinate used to compare allocation. When training samples a different monotone coordinate \(s\), with \(s\sim\pi_s\) and \(u=g(s)\), the induced density over \(u\) is
\begin{equation}
\pi(u)
=
\pi_s(s(u))
\left|
\frac{\dd s}{\dd u}
\right|.
\label{eq:path_pushforward_density}
\end{equation}
A monotone change of coordinate on a fixed path leaves the corrupted observations and posterior denoising problems unchanged, but changes the training density induced over them. After rewriting the loss in an \(x\)-prediction view, define the expected denoising loss at path location \(u\) as
\begin{equation}
R_x(u;\theta)
=
\E_{\rvx_0\sim p_{\mathrm{data}},\;\bepsilon\sim\mathcal N(\mathbf 0,\I)}
\left[
\norm{\rvx_0-\xhat(\rvx_u;u)}_2^2
\right].
\label{eq:pointwise_xrisk}
\end{equation}
Following the design-space view of \citet{karras2022elucidating} and the weighted-objective view of \citet{kingma2023understanding}, common diffusion losses can be written as weighted denoising objectives after conversion to a common prediction space. The training density \(\pi(u)\) and the induced per-noise weight \(w(u)\ge0\) therefore jointly determine the effective allocation. The conversions used in our baselines are given in \cref{app:schedule_unification}. The weighted objective is
\begin{equation}
\mathcal L(\theta)
=
\int
\underbrace{\pi(u)\,w(u)}_{\phi(u)}
\,R_x(u;\theta)\,\dd u .
\label{eq:weighted_path_objective}
\end{equation}
Here \(\pi(u)\) determines how often \(u\) is sampled, while \(w(u)\) sets the objective weight of the expected denoising loss at that location. We call \(\phi(u)\coloneqq\pi(u)w(u)\) the unnormalized effective allocation. After normalization, \(\phi\) gives the allocation profile over denoising problems.

\subsection{Information profile}
\label{sec:allocation_signal}

The effective allocation \(\phi\) specifies where the objective places training weight along the corruption path. The target for this allocation is determined by the posterior family \(p(\rvx_0\mid \rvx_u)\). Its residual uncertainty is
\begin{equation}
\Hent[\rvx_0\mid \rvx_u]
\;\defeq\;
-\E_{p(\rvx_0,\rvx_u)}
\left[
\log p(\rvx_0\mid \rvx_u)
\right].
\label{eq:condH_def_general}
\end{equation}
Since \(\Hent[\rvx_0]\) is fixed along the path,
\begin{equation}
I(\rvx_0;\rvx_u)
=
\Hent[\rvx_0]
-
\Hent[\rvx_0\mid \rvx_u],
\label{eq:mutual_information_cond_entropy}
\end{equation}
and therefore
\begin{equation}
\frac{\dd}{\dd u}
\Hent[\rvx_0\mid \rvx_u]
=
-
\frac{\dd}{\dd u}
I(\rvx_0;\rvx_u).
\label{eq:cond_entropy_mi_rate}
\end{equation}
The target is a pathwise rate, not an uncertainty level. It marks where the forward path loses mutual information and where the reverse model must reconstruct it.

At path location \(u\), the MSE-optimal \(x\)-prediction denoiser is the posterior mean \(\E[\rvx_0\mid\rvx_u]\). The corresponding unweighted Bayes risk is
\begin{equation}
\mmse(u)
\;\defeq\;
\E\!\left[
\left\|
\rvx_0-\E[\rvx_0\mid \rvx_u]
\right\|_2^2
\right]
=
\E\!\left[
\operatorname{tr}\operatorname{Cov}(\rvx_0\mid \rvx_u)
\right].
\label{eq:mmse_def_general}
\end{equation}
MMSE measures residual posterior uncertainty at a fixed path location. The allocation target is not this uncertainty level, but its rate of change along the path. Regions with small entropy rate are weak allocation targets, either because the posterior is already resolved or because neighboring path locations remain similarly unresolved.

For affine-Gaussian paths, standardizing the channel gives
\begin{equation}
\frac{\rvx_u}{b(u)}
=
\sqrt{\gamma(u)}\,\rvx_0+\bepsilon,
\qquad
\gamma(u)=\frac{a(u)^2}{b(u)^2}.
\label{eq:standardized_channel}
\end{equation}
The Gaussian-channel I--MMSE identity~\citep{guo2005mutual,palomar2005gradient} then yields
\begin{equation}
\frac{\dd}{\dd u}
\Hent[\rvx_0\mid \rvx_u]
=
-\frac12\,\gamma'(u)\,\mmse(u).
\label{eq:general_entropy_rate_identity}
\end{equation}
With \(u\) ordered from data to noise, \(\gamma'(u)\le0\). We define the information profile \(\rho^\star(u)\) as the nonnegative entropy rate
\begin{equation}
\rho^\star(u)
\;\propto\;
\frac{\dd}{\dd u}
\Hent[\rvx_0\mid \rvx_u]
=
-\frac12
\gamma'(u)
\mmse(u).
\label{eq:general_target_allocation}
\end{equation}
The factor \(|\gamma'(u)|\) converts Bayes denoising difficulty into a rate in the chosen coordinate. The profile is therefore neither raw training loss nor MMSE alone. It identifies where posterior uncertainty changes fastest along the corruption path, and gives the principled target for the effective allocation \(\phi\). This profile is the Bayes-limit counterpart of the SNR-rate denoising integrand in continuous-time VDM~\citep{kingma2021variational}. The full rewriting is given in \cref{app:vdm_entropy_rate}.

\begin{wrapfigure}[15]{r}{0.46\linewidth}
  \centering
  \includegraphics[width=\linewidth]{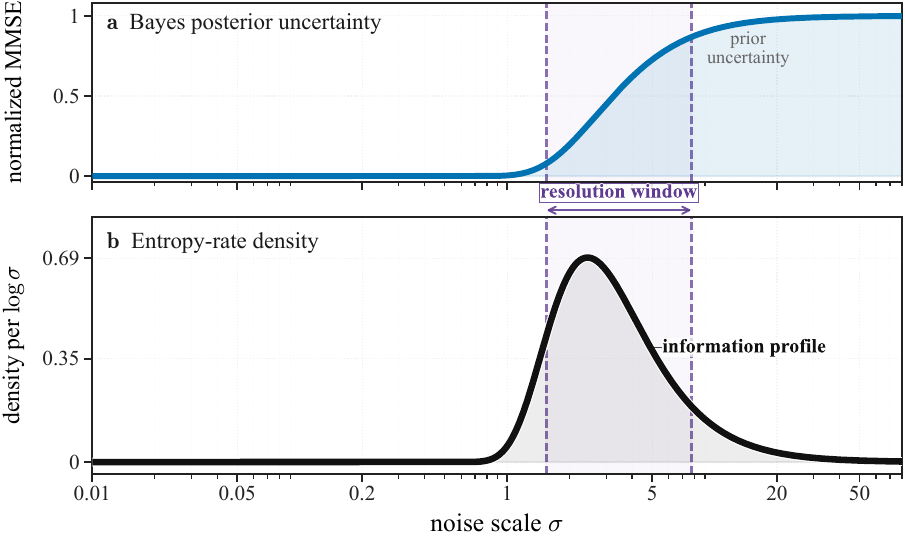}
  \caption{
  \textbf{Entropy rate localizes uncertainty resolution.}
  MMSE measures residual posterior uncertainty; entropy rate identifies where that uncertainty changes along the path.
  }
  \label{fig:toy_decision_window}
  \vspace{-.2em}
\end{wrapfigure}

\Cref{fig:toy_decision_window} shows this distinction in a symmetric two-point VE channel. MMSE tracks residual posterior uncertainty, while the entropy-rate profile localizes where that uncertainty changes. Details are given in \cref{app:toy_symmetry}.

Noise schedule design can therefore be stated as alignment between the effective allocation and the information profile. For fixed \(w\), changing the sampling density \(\pi\) changes the induced allocation \(\phi(u)=\pi(u)w(u)\). Both \(\phi\) and \(\rho^\star\) are coordinate-dependent profiles. After normalization, they transform as densities under monotone reparameterizations, so alignment is meaningful only in a common coordinate.

\paragraph{VE specialization.}
For VE paths, choosing \(u=\sigma\) gives
\(\rvx_\sigma=\rvx_0+\sigma\bepsilon\) and
\(\gamma(\sigma)=\sigma^{-2}\). Hence,
\begin{equation}
\frac{\dd}{\dd \sigma}
\Hent[\rvx_0\mid \rvx_\sigma]
=
\frac{\mmse(\sigma)}{\sigma^3}
\;\ge\;0.
\label{eq:entropy_rate_sigma}
\end{equation}
The VE form makes the coordinate dependence explicit. MMSE measures how much posterior uncertainty remains at noise level \(\sigma\), while \(\mmse(\sigma)/\sigma^3\) measures where that uncertainty is being resolved along the VE path. Offline profiles provide diagnostic comparisons of allocation. Expressions for VE, VP, sub-VP, and affine-Gaussian flow-matching paths are summarized in \cref{tab:app_schedule_summary} and derived in \cref{app:schedule_unification}.
\section{\method{} as online information-guided allocation}
\label{sec:method}

The previous section identifies the information profile as the target for effective allocation. This target depends on the Bayes MMSE and is not available before training. \method{} estimates the profile online from unweighted denoising losses and changes only the training noise distribution. Since the induced allocation is \(\phi(u)=\pi(u)w(u)\), the training noise distribution is updated so that \(\pi(u)w(u)\) follows the estimated profile. The objective, loss weighting, prediction parameterization, architecture, and optimizer remain fixed.

\subsection{Online estimation of the information profile}
\label{sec:method_target}

From \cref{sec:allocation_signal}, the target profile satisfies
\[
\rho^\star(u)
\propto
\frac12|\gamma'(u)|\mmse(u).
\]
The Bayes MMSE is unavailable during training. \method{} maintains a smoothed binwise estimate \(\hat m(u)\) from the unweighted \(x\)-space denoising losses already computed in each batch. The raw profile estimate is
\begin{equation}
\hat q(u)
\;\propto\;
\frac12\,|\gamma'(u)|\,\hat m(u),
\qquad
\hat q(\sigma)\propto \frac{\hat m(\sigma)}{\sigma^3}
\quad\text{for VE.}
\label{eq:qhat_u}
\end{equation}
The unweighted loss provides an online statistic for the denoising-error term, while the path factor converts this statistic into an entropy-rate estimate.

Before normalization, the estimate is gated near the low-noise endpoint in the training coordinate,
\begin{equation}
r(u)=\hat q(u)\,g_{c,n}(u),
\qquad
g_{c,n}(u)=\frac{u^n}{u^n+c^n},
\qquad
\hat\rho_u(u)=\frac{r(u)}{\int r(s)\,\dd s}.
\label{eq:regularized_target}
\end{equation}
The gate prevents the path factor from amplifying unstable endpoint estimates during early training, while leaving the interior profile unchanged for \(u\gg c\). Smoothing and periodic refreshes further stabilize the estimate.

\subsection{Converting the target profile into a sampler}
\label{sec:method_sampling_density}

\begin{wrapfigure}[11]{r}{0.476\linewidth}
  \centering
  \vspace{-3.5em}
  \includegraphics[width=\linewidth]{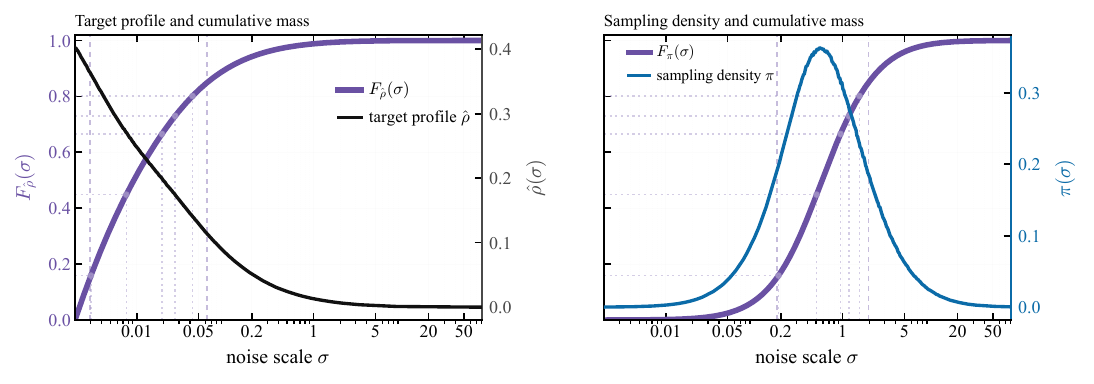}
\caption{
\textbf{Inverse-CDF sampling from the adapted density.}
Shown on CIFAR-10 for clarity, the adapted density \(\pi\) defines a cumulative mass \(F_\pi\). Uniform samples are mapped through \(F_\pi^{-1}\) to draw training noise levels. The same construction applies unchanged in other domains.
}
  \label{fig:infonoise_inverse_cdf_sampling}
  \vspace{-0.6em}
\end{wrapfigure}
The loss weight \(w(u)\) is fixed by the training objective. Matching the estimated profile requires the induced allocation to satisfy \(\pi(u)w(u)\propto\hat\rho_u(u)\), which gives
\begin{equation}
\pi(u)
=
\frac{\hat\rho_u(u)/w(u)}
{\int \hat\rho_u(s)/w(s)\,\dd s},
\qquad w(u)>0.
\label{eq:pi_from_rho_u}
\end{equation}
The resulting allocation satisfies \(\phi(u)=\pi(u)w(u)\propto\hat\rho_u(u)\). \method{} does not reweight the loss. It changes how often the fixed weighted objective visits each denoising problem.
\begin{figure*}[t]
  \centering
  \includegraphics[width=\linewidth]{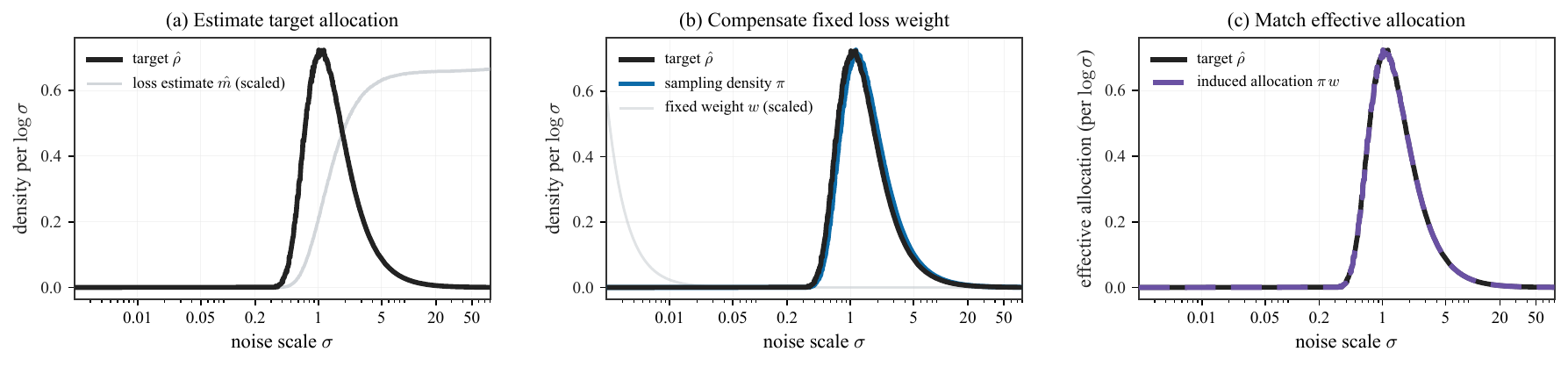}
\caption{
\textbf{\method{} converts an online information estimate into a training noise distribution.}
\textbf{(A)} Unweighted losses track \(x\)-space denoising error, which the path factor converts into an online information-profile estimate.
\textbf{(B)} With fixed loss weights, \method{} samples from \(\pi \propto \hat\rho/w\), not directly from \(\hat\rho\).
\textbf{(C)} The induced allocation \(\pi w\) matches the profile while the objective and weighting remain unchanged.
Noise levels are drawn from \(\pi\) by inverse-CDF sampling, illustrated in \cref{fig:infonoise_inverse_cdf_sampling}.
}
  \label{fig:infonoise_allocation_conversion}
\end{figure*}
\Cref{fig:infonoise_allocation_conversion} summarizes the conversion from online loss statistics to target profile, sampling density, and induced allocation.

Sampling from \(\pi\) uses
\begin{equation}
F_\pi(u)=\int_{u_{\min}}^{u}\pi(s)\,\dd s,
\qquad
\xi\sim\mathrm{Uniform}(0,1),
\qquad
u=F_\pi^{-1}(\xi).
\label{eq:inverse_cdf_sampling}
\end{equation}
In practice, \(F_\pi\) is tabulated on a fixed grid and inverted by interpolation, as illustrated in \cref{fig:infonoise_inverse_cdf_sampling}.

\newpage
\subsection{Online adaptation loop}
\label{sec:method_estimation}

\begin{wrapfigure}{r}{0.5\linewidth}
\vspace{-1.5em}
\begin{minipage}{\linewidth}
\setlength{\intextsep}{0pt}
\setlength{\algomargin}{0.7em}
\small
\begin{algorithm}[H]
\caption{\method{} online adaptation}
\label{alg:infonoise_training}
Initialize \(\hat m(u)\equiv 1\), choose warm-up prior \(\pi_0(u)\), set \(\pi\gets\pi_0\)\;
\For{training steps}{
  Sample \(\rvx_0\sim p_{\mathrm{data}}\), \(u\sim\pi(u)\), \(\bepsilon\sim\mathcal N(\mathbf 0,\I)\), and set \(\rvx_u=a(u)\rvx_0+b(u)\bepsilon\)\;
  Compute \(\ell=\|\rvx_0-\xhat(\rvx_u;u)\|_2^2\), update model with \(w(u)\ell\), and update \(\hat m\) with \((u,\ell)\)\;
  \If{refresh step}{
    Smooth \(\hat m\), set \(\hat q(u)\gets \frac12|\gamma'(u)|\hat m(u)\), gate and normalize to obtain \(\hat\rho_u\), then update \(\pi(u)\propto\hat\rho_u(u)/w(u)\)\;
  }
}
\end{algorithm}
\end{minipage}
\vspace{-1.0em}
\end{wrapfigure}

\Cref{alg:infonoise_training} summarizes the online update. The estimator is maintained on a fixed grid in the training coordinate \(u\). Each batch contributes the unweighted denoising loss \(\ell\) to the profile estimate, while the model update still uses the original weighted loss \(w(u)\ell\). The estimator changes the sampler, not the objective.

Warm-up uses a fixed prior \(\pi_0\) to collect initial loss statistics. After the first refresh, the sampler is rebuilt from \(\hat\rho_u(u)/w(u)\), so the induced allocation follows the learned profile rather than the warm-up prior. The estimator, gate, smoothing rule, and refresh cadence remain fixed. Only the sampling distribution adapts online.
\section{Experiments}
\label{sec:experiments}

We now present empirical evidence that the information profile can serve as an automatic target for noise-schedule design. Standard image diffusion provides the already-optimized regime. Schedules such as EDM and cosine have been refined through extensive practice, so their allocations provide strong fixed references. In this regime, adaptation should preserve matched-budget quality rather than disturb an already useful allocation. The harder case is transfer. When the denoising problem changes through representation, sequence structure, or modality, inherited schedules need not place training effort near the informative region. We therefore evaluate mature image benchmarks as optimized regimes, and binarized images, digitized images, DNA regulatory sequences following the DNA-Diffusion setup~\citep{dasilva2026designing}, and text as shifted regimes. The expected pattern is asymmetric. Information-guided adaptation should preserve matched-budget quality when inherited allocations are aligned and reduce training compute when the informative region moves.

\textbf{Experimental design.}
All comparisons isolate training allocation. Schedules are expressed in the common \(x\)-prediction view, where sampling density and loss weight induce an effective allocation over denoising problems~\citep{kingma2023understanding}. Within each domain, only the training noise distribution varies. The model, objective, loss weighting, optimizer, EMA, augmentations, inference sampler, and evaluation budget are fixed. Fixed baselines include image-tuned EDM sampling~\citep{karras2022elucidating}, variance-preserving cosine sampling~\citep{nichol2021improved}, flow-matching OT sampling~\citep{lipman2023flow}, and log-uniform sampling in \(\sigma\). KG-adaptive is the adaptive control, following weighted-loss magnitude rather than the entropy-rate profile~\citep{kingma2023understanding}. Offline profiles are used only for alignment analysis and are never provided to \method{} during training. Dataset-specific metrics are reported in \cref{tab:main_results_quality_efficiency}. Schedule conversions are given in \cref{app:schedule_unification}; architectures, metrics, training details, and settings are given in \cref{app:experimental_details}.

\subsection{Information-guided allocation preserves tuned regimes and accelerates transfer}
\label{sec:exp_main_results}

\begin{figure}
    \centering
    \includegraphics[width=\linewidth]{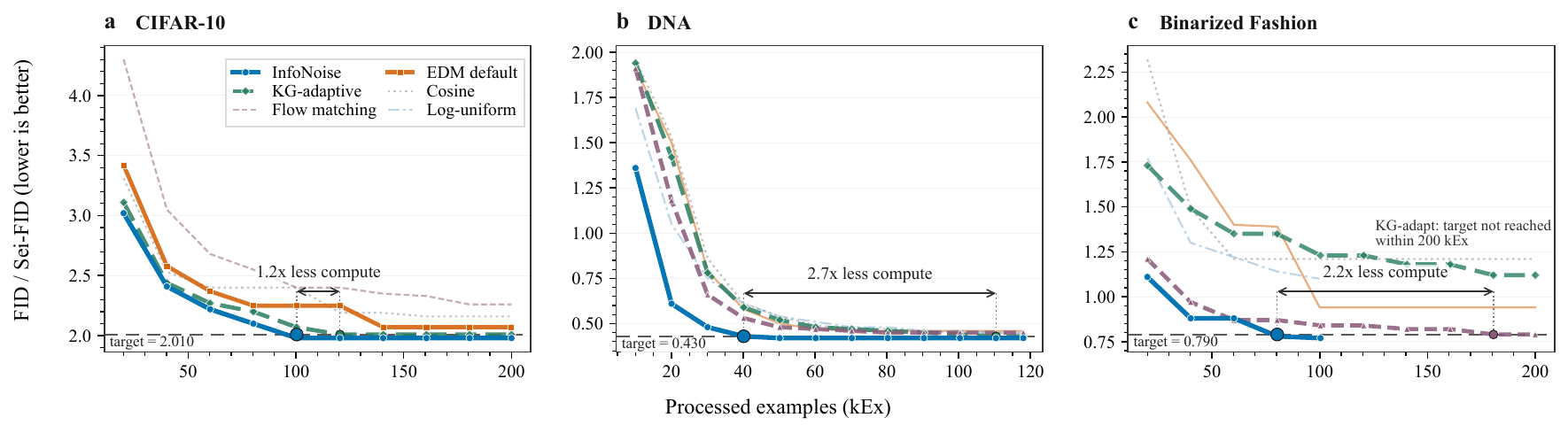} 
    \caption{
    \textbf{\method{} reaches the best baseline quality with fewer training updates.}
    For each dataset, the target is the final matched-budget quality of the strongest non-\(\method{}\) baseline, including fixed schedules and KG-adaptive. Speedup reports how much less training compute \method{} needs to reach that target. A value of \(1.00\) means no measured reduction within the training budget; values above \(1.00\) indicate that the same quality is reached with fewer updates.
    } 
  \label{fig:matched_shifted_speedup}
\end{figure}
\begin{table*}[!t]
  \centering
  \scriptsize
  \setlength{\tabcolsep}{3.0pt}
  \renewcommand{\arraystretch}{1.08}
  \resizebox{\linewidth}{!}{%
  \begin{tabular}{@{}llc
    S[table-format=2.2]
    S[table-format=2.2]
    S[table-format=2.2]
    S[table-format=2.2]
    S[table-format=2.2]
    >{\columncolor{methodbg}}S[table-format=2.2]
    S[table-format=1.2]@{}}
    \toprule
    & & &
    \multicolumn{4}{c}{\textbf{Fixed schedules}} &
    \multicolumn{2}{c}{\textbf{Adaptive schedules}} &
    \multicolumn{1}{c}{\textbf{Efficiency}} \\
    \cmidrule(lr){4-7}
    \cmidrule(lr){8-9}
    \cmidrule(l){10-10}
    Dataset
    & Metric
    & NFE
    & \multicolumn{1}{c}{Log-unif.}
    & \multicolumn{1}{c}{Cosine}
    & \multicolumn{1}{c}{FM-OT}
    & \multicolumn{1}{c}{EDM}
    & \multicolumn{1}{c}{KG-adapt.}
    & \multicolumn{1}{c}{\textbf{\method{}}}
    & \multicolumn{1}{c}{\(\times\) vs fixed} \\
    \midrule

    MNIST
      & FID $\downarrow$
      & 35
      & 1.02
      & 0.46
      & 0.44
      & 0.44
      & \multicolumn{1}{c}{\bfseries 0.39}
      & 0.41
      & \bfseries1.20 \\

    FashionMNIST
      & FID $\downarrow$
      & 35
      & 2.75
      & 1.76
      & 1.93
      & 1.78
      & 1.82
      & \multicolumn{1}{c}{\cellcolor{methodbg}\bfseries 1.71}
      & \multicolumn{1}{c}{\bfseries 1.45} \\

    CIFAR-10 uncond.
      & FID $\downarrow$
      & 35
      & 5.93
      & 2.16
      & 2.31
      & 2.04
      & 2.00
      & \multicolumn{1}{c}{\cellcolor{methodbg}\bfseries 1.98}
      & \multicolumn{1}{c}{\bfseries 1.40} \\

    CIFAR-10 cond.
      & FID $\downarrow$
      & 35
      & 4.45
      & 2.11
      & 2.13
      & 1.85
      & \multicolumn{1}{c}{\bfseries 1.84}
      & \multicolumn{1}{c}{\cellcolor{methodbg}\bfseries 1.84}
      & \multicolumn{1}{c}{\bfseries 1.50} \\

    FFHQ \(64{\times}64\)
      & FID $\downarrow$
      & 79
      & 4.68
      & 2.69
      & 2.72
      & \multicolumn{1}{c}{\bfseries 2.53}
      & 2.54
      & \multicolumn{1}{c}{\cellcolor{methodbg}\bfseries 2.53}
      & \multicolumn{1}{c}{\bfseries 1.00} \\

    bMNIST
      & FID $\downarrow$
      & 35
      & 0.63
      & 0.52
      & 0.38
      & 0.53
      & 0.38
      & \multicolumn{1}{c}{\cellcolor{methodbg}\bfseries 0.33}
      & \multicolumn{1}{c}{\bfseries 2.20} \\

    bFashionMNIST
      & FID $\downarrow$
      & 35
      & 1.10
      & 1.20
      & 0.80
      & 0.94
      & 1.22
      & \multicolumn{1}{c}{\cellcolor{methodbg}\bfseries 0.69}
      & \multicolumn{1}{c}{\bfseries 3.00} \\

    Digitized CIFAR-10
      & FID $\downarrow$
      & 255
      & 17.30
      & \multicolumn{1}{c}{\textemdash}
      & \multicolumn{1}{c}{\textemdash}
      & 9.90
      & \multicolumn{1}{c}{\textemdash}
      & \multicolumn{1}{c}{\cellcolor{methodbg}\bfseries 3.10}
      & \multicolumn{1}{c}{\bfseries 3.00} \\

    DNA
      & Sei-FID $\downarrow$
      & 35
      & 0.61
      & \multicolumn{1}{c}{\bfseries 0.42}
      & 0.44
      & 0.45
      & 0.43
      & \multicolumn{1}{c}{\cellcolor{methodbg}\bfseries 0.42}
      & \multicolumn{1}{c}{\bfseries 2.00} \\

    OpenWebText-2
      & MAUVE $\uparrow$
      & 127
      & 0.68
      & \multicolumn{1}{c}{\textemdash}
      & \multicolumn{1}{c}{\textemdash}
      & 0.83
      & \multicolumn{1}{c}{\textemdash}
      & \multicolumn{1}{c}{\cellcolor{methodbg}\bfseries 0.87}
      & \multicolumn{1}{c}{\bfseries 2.06} \\

    \bottomrule
  \end{tabular}}%
\caption{
\textbf{Information-guided allocation automates schedule adaptation across regimes.}
Each row compares schedules trained to the same budget and evaluated at the listed NFE. Quality is the final metric at that budget. Efficiency reports the compute \method{} needs to reach the final metric of the best competing baseline; values above one mean the same target is reached with fewer updates. Shading denotes \method{}; bold marks best final quality and speedups above one.
}
  \label{tab:main_results_quality_efficiency}
\end{table*}
\Cref{tab:main_results_quality_efficiency,fig:matched_shifted_speedup} show that information-guided allocation can reproduce the benefit of tuned image schedules and adapt online during training when those schedules no longer match the denoising problem, without additional schedule retuning. On MNIST, FashionMNIST, CIFAR-10, and FFHQ, fixed schedules such as EDM and cosine are strong references because their allocations have been refined through empirical practice. In these optimized regimes, \method{} preserves matched-budget quality against the strongest baselines and often reaches the same target with fewer updates. The more demanding case is transfer. When representation, sequence structure, or modality changes, inherited image schedules need not place training effort near the informative region. On binarized images, digitized CIFAR-10, DNA, and OpenWebText-2, \method{} matches or improves final quality while reaching the strongest non-\(\method{}\) baseline target with up to \(3\times\) less training compute. The information profile therefore provides an online target for schedule design, replacing manual search over fixed schedule families across these settings. The next subsection tests this mechanism by comparing effective allocations with measured information profiles.

\subsection{Profile alignment explains schedule transfer}
\label{sec:exp_profile_alignment}

\begin{figure*}[t]
  \centering
  \includegraphics[width=\linewidth]{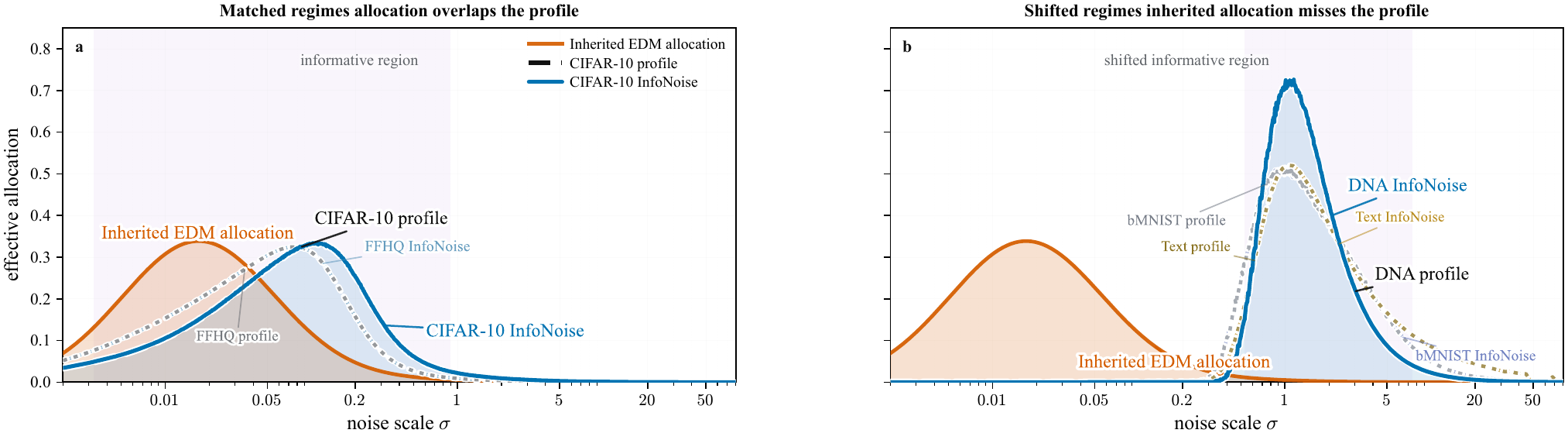}
\caption{
\textbf{Profile alignment explains schedule transfer.}
Post hoc information profiles are compared with the effective allocation induced by inherited EDM sampling and by \method{}.
\textbf{(A)} In matched image regimes, inherited allocation overlaps the informative region, and \method{} converges to a similar allocation online.
\textbf{(B)} Under shift, the information profile moves away from inherited image allocation. \method{} moves the induced allocation toward the shifted profile without schedule retuning.
Profiles are used only for analysis and never provided to \method{} during training.
}
  \label{fig:matched_shifted_alignment}
\end{figure*}

\Cref{fig:matched_shifted_alignment} compares the measured information profile with the effective allocation induced by inherited EDM sampling and by \method{}. The profiles are computed after training from reference denoisers, used only for analysis, and never provided to \method{} during training. In CIFAR-10 and FFHQ, the inherited image allocation already overlaps the informative region. \method{} recovers a similar allocation online, showing that the online information estimate can recover the region that image schedule design has found empirically. Under representation, sequence, and modality shift, this overlap no longer holds. In binarized images, DNA, and text, the fixed schedule continues to spend updates according to the image regime, while \method{} moves the induced allocation toward the shifted profile without schedule retuning. The largest speedups in \cref{tab:main_results_quality_efficiency} occur in these displaced-profile regimes. Matched image regimes therefore test automatic recovery of a tuned allocation; shifted regimes test adaptation when that allocation no longer follows the denoising difficulty.

\subsection{Schedule search is allocation search}
\label{sec:exp_schedule_retuning}

\begin{figure*}
  \centering
  \includegraphics[width=\linewidth]{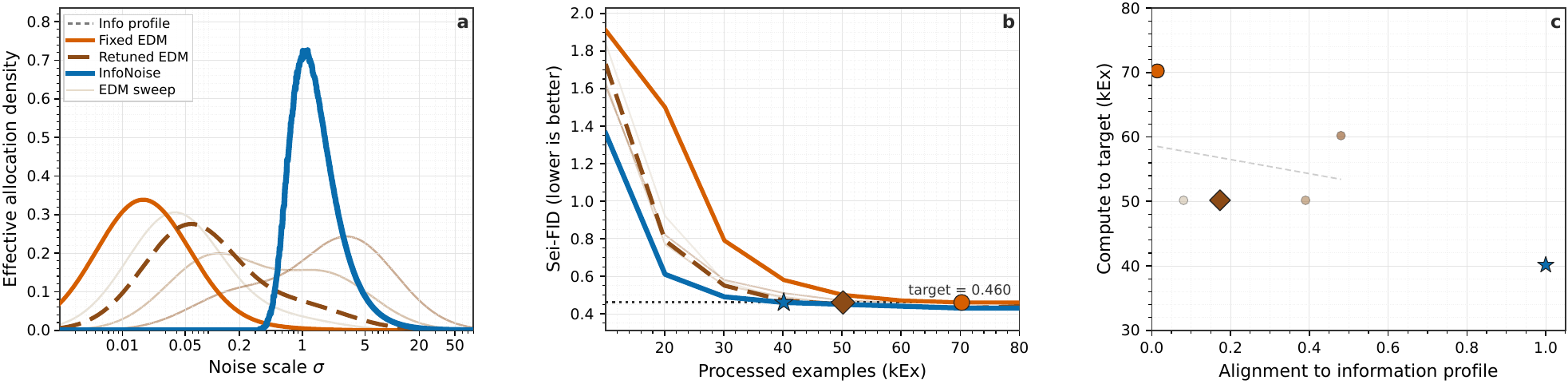}
\caption{
\textbf{Schedule search is allocation search.}
\textbf{(a)} Sweeping \(P_{\mathrm{mean}}\) in the EDM log-normal schedule shifts effective allocation along the DNA corruption path while other training choices remain fixed. \method{} reaches the informative region online in one run.
\textbf{(b)} Allocations closer to the information profile reach the strongest non-\(\method{}\) baseline target with fewer updates.
\textbf{(c)} Training compute decreases as effective allocation becomes better aligned with the DNA information profile, showing that misalignment with the profile is what makes schedule transfer expensive.
}
  \label{fig:dna_allocation_search}
\end{figure*}

When a schedule is transferred to a new domain, the usual remedy is to retune its parameters by training several candidates and selecting the best validation run. In the allocation view, this procedure is not an abstract hyperparameter search. It is a search over fixed effective allocations. The EDM log-normal training distribution provides a controlled instance of this search. Holding the objective, loss weighting, prediction parameterization, model, optimizer, sampler, and evaluation fixed, changing \(P_{\mathrm{mean}}\) in \(\log\sigma\sim\mathcal N(P_{\mathrm{mean}},P_{\mathrm{std}}^2)\), with \(P_{\mathrm{std}}\) fixed, shifts only the sampling density \(\pi\), and therefore the effective allocation \(\phi=\pi w\). DNA is a shifted regime where the inherited image allocation lies away from the measured information profile. \Cref{fig:dna_allocation_search} shows that moving \(P_{\mathrm{mean}}\) moves allocation along the DNA corruption path, and that performance improves as this allocation approaches the informative region. Settings that remain near the inherited image allocation converge more slowly and reach worse Sei-FID; settings closer to the profile reach the target with less compute. \method{} reaches the same region in a single run by adapting allocation toward the information profile, replacing repeated schedule trials with information-guided noise allocation.

\subsection{Online adaptation overrides warm-up and stabilizes}
\label{sec:exp_online}

\begin{wraptable}[12]{r}{.3\linewidth}
  \vspace{-1.2em}
  \centering
  \small
  \setlength{\tabcolsep}{4pt}
  \renewcommand{\arraystretch}{1.05}
  \begin{tabular}{@{}lc@{}}
    \toprule
    Warm-up prior & Sei-FID $\downarrow$ \\
    \midrule
    EDM log-normal & 0.460 \\
    Log-uniform in \(\sigma\) & 0.463 \\
    Log-\(\sigma\) const.-entropy & \textbf{0.456} \\
    \midrule
    Spread & \(0.007\) \\
    \bottomrule
  \end{tabular}
  \caption{
  \textbf{Warm-up does not set final performance.}
  DNA Sei-FID changes by only \(0.007\) across warm-up priors.
  }
  \label{tab:warmup_ablation}
\end{wraptable}

\begin{figure*}[t]
  \centering
  \begin{minipage}[t]{0.49\linewidth}
    \centering
    \includegraphics[width=\linewidth]{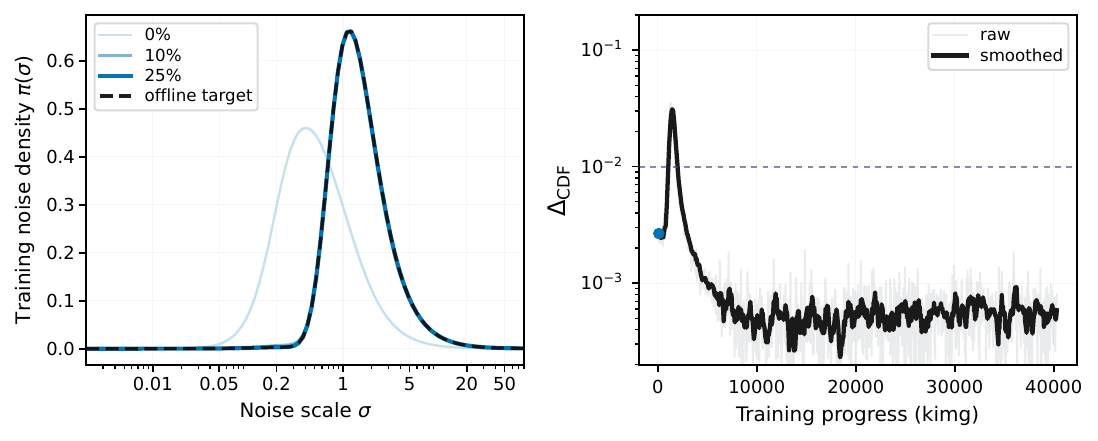}
  \end{minipage}\hfill
  \begin{minipage}[t]{0.49\linewidth}
    \centering
    \includegraphics[width=\linewidth]{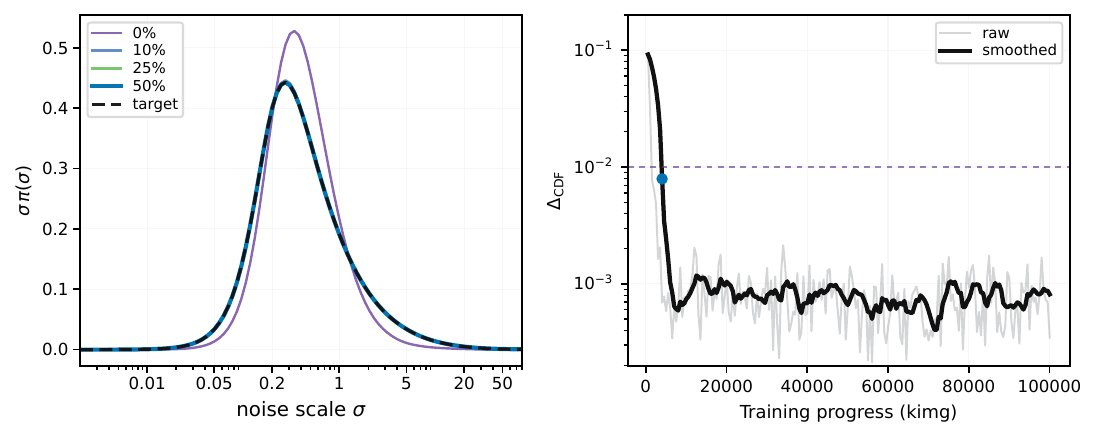}
  \end{minipage}
\caption{
\textbf{\method{} rapidly recovers the information profile and stabilizes.}
On DNA, the online estimate moves from the warm-up allocation toward the post hoc diagnostic profile within the first rebuilds. The diagnostic profile is computed after training and never provided to \method{}. Sampler-CDF changes fall below \(10^{-2}\), showing that the sampler stabilizes quickly rather than tracking transient loss fluctuations.
}
\vspace{-1em}
  \label{fig:online_region_learning}
\end{figure*}

Adaptive sampling starts from a prior only because no profile estimate exists at initialization. On DNA, \cref{fig:online_region_learning} shows that this prior is quickly superseded. The online estimate moves from the warm-up allocation toward the post hoc diagnostic information profile within the first rebuilds, while the sampler rebuilds the induced effective allocation around this estimate. The diagnostic profile is computed only after training from a reference denoiser and is never provided to \method{}. Consecutive sampler-CDF changes fall below \(10^{-2}\) after the first rebuilds, indicating that the sampler stabilizes rather than following transient loss fluctuations. \Cref{tab:warmup_ablation} gives the corresponding endpoint check. Changing only the warm-up prior across EDM, log-uniform, and log-\(\sigma\) constant-entropy starts changes final Sei-FID by \(0.007\). The DNA result is therefore driven by the online profile update after warm-up, not by the initial sampling prior.

\subsection{Training-allocation gains persist under guidance}
\label{sec:exp_cfg}

\begin{wrapfigure}[10]{r}{0.31\linewidth}
  \vspace{-5.0em}
  \centering
  \includegraphics[width=\linewidth]{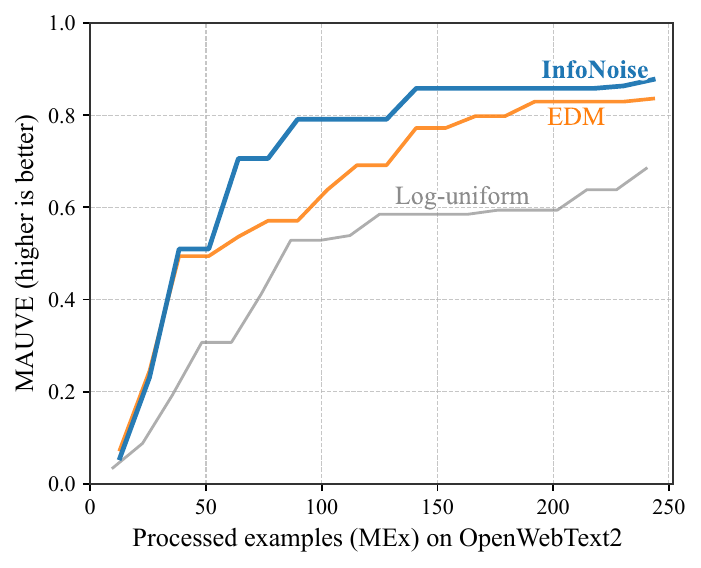}
  \caption{
  \textbf{Training-allocation gains persist under fixed CFG.}
  With guidance scale and sampler fixed, \method{} reaches higher MAUVE earlier.
  }
  \label{fig:cfg_mauve}
  \vspace{-1.0em}
\end{wrapfigure}

Classifier-free guidance changes how the trained conditional and unconditional predictions are combined at sampling time. We keep this inference procedure fixed across schedules by using the same guidance scale, sampler, and generation budget. The CFG training recipe is also fixed, including the conditional-dropout setup. Thus the only difference between runs is the training noise distribution. \Cref{fig:cfg_mauve} shows that \method{} reaches higher MAUVE earlier under the same guided sampler. The allocation learned during training therefore remains beneficial under guided inference, rather than depending on guidance-scale or sampler tuning.

\section{Related work}
\label{sec:related_work}

\textbf{Noise schedules, weighting, and training allocation.}
Much of diffusion training design can be read as deciding where denoising errors should matter along the corruption path. Noise schedules determine which path locations are visited, while loss weights and prediction parameterizations determine how their errors enter the objective. Early diffusion and score-based models fixed this choice through variance schedules, log-spaced noise scales, and noise-dependent objectives~\citep{sohl2015deep,song2019generative,ho2020denoising,song2020score}. Later work refined the allocation through cosine and learned schedules, log-SNR parameterizations, SNR-aware weighting, importance sampling, and loss-aware sampling~\citep{nichol2021improved,kingma2021variational,karras2022elucidating,kingma2023understanding,hang2023efficient,hoogeboom2023simple,chen2023importance}. Spectral approaches make schedule ranges instance-dependent for pixel diffusion by using image frequency content to remove redundant noise levels and improve low-step sampling~\citep{esteves2026spectrally}. EDM separates path, preconditioning, weighting, sampler, parameterization, and training noise distribution, but the training distribution remains chosen beforehand or retuned across runs~\citep{karras2022elucidating}. Flow matching and rectified flows expose the same dependence on path-time sampling~\citep{lipman2023flow,esser2024scaling}. Discrete and language diffusion further expose schedule-transfer limits through absorbing masks, mutual-information schedules, state-dependent masking, contextual schedulers, and schedule-conditioned transitions~\citep{austin2021structured,dieleman2022continuous,he2023diffusionbert,shi2024simplified,sahoo2024simple,chen2026langflow}. \method{} keeps the objective, parameterization, and loss weight fixed, and adapts only the training noise distribution so that the induced allocation follows an online estimate of the information profile.

\textbf{Information transfer as a schedule-design target.}
The allocation induced by a training objective needs a target. For Gaussian channels, I--MMSE links information-change rate to Bayes denoising error~\citep{guo2005mutual,palomar2005gradient}. This identity has been used for diffusion objectives, likelihoods, regularization, and path geometry~\citep{kong2023informationtheoretic,franzese2024minde,wang2023infodiffusion,premkumar2025neural,wang2025informationtheoretic,ambrogioni2025information}. Entropy-based schedulers use information signals for time reparameterization or inference-step allocation~\citep{stancevic2025entropic}, and LangFlow uses information gain for language-diffusion schedules~\citep{chen2026langflow}. \method{} brings the conditional-entropy-rate profile to training, estimates it online from denoising losses, and uses it as the target for the effective allocation induced by sampling and weighting.

\textbf{Localized denoising difficulty and critical regions.}
Denoising difficulty is not uniform along the generative path. Target-field variance can concentrate in restricted regions, motivating modified targets, reference information, or changes to the learned field~\citep{xu2023stable,yang2026stable}. Work on symmetry breaking and critical dynamics shows a complementary form of localization. Diffusion trajectories do not recover structure gradually and uniformly. They pass through windows where symmetries break, commitments form, and sample structure emerges over a restricted range of noise levels~\citep{raya2024spontaneous,biroli2024dynamical,li2024critical,sclocchi2025phase}. These results make noise level choice part of the mechanism of pattern formation, not only a numerical discretization choice. In our allocation view, this localization appears as a profile of uncertainty resolution along the corruption path. \method{} uses this profile as a training-side allocation target while keeping the model, objective, parameterization, and sampler fixed.
\section{Discussion and conclusions}
\label{sec:discussion_conclusion}

Diffusion training allocates optimization effort across denoising problems. The allocation view puts noise schedules and loss weights into one object, the effective allocation along the corruption path. This reframes noise-schedule design as deciding where training effort should be placed, rather than selecting a fixed schedule family inherited from another regime. The information profile provides the target. For affine-Gaussian corruption paths, this profile is the conditional-entropy rate, linked by I--MMSE to Bayes denoising difficulty. \method{} estimates this profile online and adapts only the training noise distribution, while leaving the objective, loss weighting, parameterization, architecture, optimizer, and sampler fixed. The empirical results follow this structure. When inherited allocations match the profile, as in optimized image regimes, \method{} preserves matched-budget quality. When the profile shifts under representation, sequence, or modality change, information-guided adaptation redirects allocation and reaches strong baseline targets with less compute. The alignment, schedule-search, online-recovery, and warm-up analyses support a single interpretation. The gains come from adapting effective allocation toward the information profile, not from loss-magnitude resampling, favorable initialization, or post hoc access to the profile. Noise-schedule design should therefore move from retuning inherited schedule families to estimating the task-specific information structure of the denoising problem. Fixed schedules remain useful when their allocation matches this structure. When the profile moves, schedule search becomes an indirect way of rediscovering where uncertainty is resolved. \method{} makes that structure available during training and turns schedule design into online information-guided allocation matched to the task-specific denoising problem.

\textbf{Limitations and future work.}

The experiments support a direct view of schedule design. Efficient schedules place effective allocation near the noise levels where posterior uncertainty changes fastest. The main remaining limitation is boundary behavior. Conditional-entropy-rate estimates can become singular near deterministic limits, so a more principled regularization could sharpen the profile estimate and improve endpoint accuracy. We did not test high-resolution regimes, where the profile may decompose across scales or semantic levels. Extending the allocation view to structured, blockwise, or multimodal corruption paths is a natural next step, since components may resolve uncertainty at different stages of the generative trajectory.

\section*{Impact Statement}
\label{sec:impact_statement}

This paper advances the methodological foundations of  diffusion training by replacing manual noise-schedule search with a data-adaptive allocation principle grounded in information dynamics along the corruption path. Reducing trial-and-error schedule design can lower the compute and engineering burden of training diffusion models across new datasets, resolutions, and representations, accelerating research and broadening access. Improved efficiency can also lower the marginal cost of high-fidelity synthetic data, amplifying both beneficial applications and harmful misuse.

\section*{Acknowledgements}
Gabriel Raya was funded by the Dutch Research Council (NWO) as part of the CERTIF-AI project (file number 17998). 
\bibliographystyle{apalike}
\bibliography{ref}

@inproceedings{hang2023efficient,
  title={Efficient diffusion training via min-snr weighting strategy},
  author={Hang, Tiankai and Gu, Shuyang and Li, Chen and Bao, Jianmin and Chen, Dong and Hu, Han and Geng, Xin and Guo, Baining},
  booktitle={Proceedings of the IEEE/CVF international conference on computer vision},
  pages={7441--7451},
  year={2023}
}

@article{lai2025principles,
  title={The principles of diffusion models},
  author={Lai, Chieh-Hsin and Song, Yang and Kim, Dongjun and Mitsufuji, Yuki and Ermon, Stefano},
  journal={arXiv preprint arXiv:2510.21890},
  year={2025}
}

@article{chen2026langflow,
  title={LangFlow: Continuous Diffusion Rivals Discrete in Language Modeling},
  author={Chen, Yuxin and Liang, Chumeng and Sui, Hangke and Guo, Ruihan and Cheng, Chaoran and You, Jiaxuan and Liu, Ge},
  journal={arXiv preprint arXiv:2604.11748},
  year={2026}
}

@inproceedings{sohl2015deep,
  title={Deep unsupervised learning using nonequilibrium thermodynamics},
  author={Sohl-Dickstein, Jascha and Weiss, Eric and Maheswaranathan, Niru and Ganguli, Surya},
  booktitle={International Conference on Machine Learning},
  year={2015}
}

@article{ho2020denoising,
  title={Denoising diffusion probabilistic models},
  author={Ho, Jonathan and Jain, Ajay and Abbeel, Pieter},
  journal={Advances in neural information processing systems},
  volume={33},
  pages={6840--6851},
  year={2020}
}

@article{song2019generative,
  title={Generative modeling by estimating gradients of the data distribution},
  author={Song, Yang and Ermon, Stefano},
  journal={Advances in neural information processing systems},
  volume={32},
  year={2019}
}

@article{song2020score,
  title={Score-based generative modeling through stochastic differential equations},
  author={Song, Yang and Sohl-Dickstein, Jascha and Kingma, Diederik P and Kumar, Abhishek and Ermon, Stefano and Poole, Ben},
  journal={arXiv preprint arXiv:2011.13456},
  year={2020}
}

@inproceedings{nichol2021improved,
  title={Improved denoising diffusion probabilistic models},
  author={Nichol, Alexander Quinn and Dhariwal, Prafulla},
  booktitle={International conference on machine learning},
  pages={8162--8171},
  year={2021},
  organization={PMLR}
}

@article{karras2022elucidating,
  title={Elucidating the design space of diffusion-based generative models},
  author={Karras, Tero and Aittala, Miika and Aila, Timo and Laine, Samuli},
  journal={Advances in neural information processing systems},
  volume={35},
  pages={26565--26577},
  year={2022}
}

@article{li2024critical,
  title={Critical windows: non-asymptotic theory for feature emergence in diffusion models},
  author={Li, Marvin and Chen, Sitan},
  journal={arXiv preprint arXiv:2403.01633},
  year={2024}
}

@article{kingma2021variational,
  title={Variational diffusion models},
  author={Kingma, Diederik and Salimans, Tim and Poole, Ben and Ho, Jonathan},
  journal={Advances in neural information processing systems},
  volume={34},
  pages={21696--21707},
  year={2021}
}

@inproceedings{karras2024analyzing,
  title={Analyzing and improving the training dynamics of diffusion models},
  author={Karras, Tero and Aittala, Miika and Lehtinen, Jaakko and Hellsten, Janne and Aila, Timo and Laine, Samuli},
  booktitle={Proceedings of the IEEE/CVF conference on computer vision and pattern recognition},
  pages={24174--24184},
  year={2024}
}

@article{biroli2024dynamical,
  title={Dynamical regimes of diffusion models},
  author={Biroli, Giulio and Bonnaire, Tony and De Bortoli, Valentin and M{\'e}zard, Marc},
  journal={Nature Communications},
  volume={15},
  number={1},
  pages={9957},
  year={2024},
  publisher={Nature Publishing Group UK London}
}

@article{raya2024spontaneous,
  title={Spontaneous symmetry breaking in generative diffusion models},
  author={Raya, Gabriel and Ambrogioni, Luca},
  journal={Advances in Neural Information Processing Systems},
  volume={36},
  pages={66377--66389},
  year={2023}
}

@article{stancevic2025entropic,
  title={Entropic time schedulers for generative diffusion models},
  author={Stancevic, Dejan and Handke, Florian and Ambrogioni, Luca},
  journal={Advances in Neural Information Processing Systems},
  volume={38},
  pages={44222--44252},
  year={2026}
}

@inproceedings{lipman2023flow,
title={Flow Matching for Generative Modeling},
author={Yaron Lipman and Ricky T. Q. Chen and Heli Ben-Hamu and Maximilian Nickel and Matthew Le},
booktitle={The Eleventh International Conference on Learning Representations },
year={2023},
url={https://openreview.net/forum?id=PqvMRDCJT9t}
}

@article{chen2023importance,
  title={On the importance of noise scheduling for diffusion models},
  author={Chen, Ting},
  journal={arXiv preprint arXiv:2301.10972},
  year={2023}
}

@article{kingma2023understanding,
  title={Understanding diffusion objectives as the elbo with simple data augmentation},
  author={Kingma, Diederik and Gao, Ruiqi},
  journal={Advances in Neural Information Processing Systems},
  volume={36},
  pages={65484--65516},
  year={2023}
}

@Techreport{krizhevsky2009learning,
 author = {Krizhevsky, Alex and Hinton, Geoffrey},
 address = {Toronto, Ontario},
 institution = {University of Toronto},
 number = {0},
 publisher = {Technical report, University of Toronto},
 title = {Learning multiple layers of features from tiny images},
 year = {2009},
 title_with_no_special_chars = {Learning multiple layers of features from tiny images},
 url = {https://www.cs.toronto.edu/~kriz/learning-features-2009-TR.pdf}
}

@inproceedings{karras2019style,
  title={A style-based generator architecture for generative adversarial networks},
  author={Karras, Tero and Laine, Samuli and Aila, Timo},
  booktitle={Proceedings of the IEEE/CVF conference on computer vision and pattern recognition},
  pages={4401--4410},
  year={2019}
}

@Article{ambrogioni2025information,
AUTHOR = {Stančević, Dejan and Ambrogioni, Luca},
TITLE = {The Information Dynamics of Generative Diffusion},
JOURNAL = {Entropy},
VOLUME = {28},
YEAR = {2026},
NUMBER = {2},
ARTICLE-NUMBER = {195},
URL = {https://www.mdpi.com/1099-4300/28/2/195},
PubMedID = {41751698},
ISSN = {1099-4300},
DOI = {10.3390/e28020195}
}

@inproceedings{rombach2022high,
  title={High-resolution image synthesis with latent diffusion models},
  author={Rombach, Robin and Blattmann, Andreas and Lorenz, Dominik and Esser, Patrick and Ommer, Bj{\"o}rn},
  booktitle={Proceedings of the IEEE/CVF conference on computer vision and pattern recognition},
  pages={10684--10695},
  year={2022}
}

@article{kong2020diffwave,
  title={Diffwave: A versatile diffusion model for audio synthesis},
  author={Kong, Zhifeng and Ping, Wei and Huang, Jiaji and Zhao, Kexin and Catanzaro, Bryan},
  journal={arXiv preprint arXiv:2009.09761},
  year={2020}
}

@article{ho2022video,
  title={Video diffusion models},
  author={Ho, Jonathan and Salimans, Tim and Gritsenko, Alexey and Chan, William and Norouzi, Mohammad and Fleet, David J},
  journal={Advances in neural information processing systems},
  volume={35},
  pages={8633--8646},
  year={2022}
}

@article{dieleman2022continuous,
  title={Continuous diffusion for categorical data},
  author={Dieleman, Sander and Sartran, Laurent and Roshannai, Arman and Savinov, Nikolay and Ganin, Yaroslav and Richemond, Pierre H and Doucet, Arnaud and Strudel, Robin and Dyer, Chris and Durkan, Conor and others},
  journal={arXiv preprint arXiv:2211.15089},
  year={2022}
}

@inproceedings{hoogeboom2023simple,
  title={simple diffusion: End-to-end diffusion for high resolution images},
  author={Hoogeboom, Emiel and Heek, Jonathan and Salimans, Tim},
  booktitle={International Conference on Machine Learning},
  pages={13213--13232},
  year={2023},
  organization={PMLR}
}

@article{guo2005mutual,
  title={Mutual information and minimum mean-square error in Gaussian channels},
  author={Guo, Dongning and Shamai, Shlomo and Verd{\'u}, Sergio},
  journal={IEEE transactions on information theory},
  volume={51},
  number={4},
  pages={1261--1282},
  year={2005},
  publisher={IEEE}
}

@inproceedings{esser2024scaling,
  title={Scaling rectified flow transformers for high-resolution image synthesis},
  author={Esser, Patrick and Kulal, Sumith and Blattmann, Andreas and Entezari, Rahim and M{\"u}ller, Jonas and Saini, Harry and Levi, Yam and Lorenz, Dominik and Sauer, Axel and Boesel, Frederic and others},
  booktitle={Forty-first international conference on machine learning},
  year={2024}
}

@article{palomar2005gradient,
  author={Palomar, D.P. and Verdu, S.},
  journal={IEEE Transactions on Information Theory}, 
  title={Gradient of mutual information in linear vector Gaussian channels}, 
  year={2006},
  volume={52},
  number={1},
  pages={141-154},
  doi={10.1109/TIT.2005.860424}
}

@article{austin2021structured,
  title={Structured denoising diffusion models in discrete state-spaces},
  author={Austin, Jacob and Johnson, Daniel D and Ho, Jonathan and Tarlow, Daniel and Van Den Berg, Rianne},
  journal={Advances in neural information processing systems},
  volume={34},
  pages={17981--17993},
  year={2021}
}

@inproceedings{
    kong2023informationtheoretic,
    title={Information-Theoretic Diffusion},
    author={Xianghao Kong and Rob Brekelmans and Greg Ver Steeg},
    booktitle={The Eleventh International Conference on Learning Representations },
    year={2023},
    url={https://openreview.net/forum?id=UvmDCdSPDOW}
}

@inproceedings{franzese2024minde,
  title={MINDE: Mutual information neural diffusion estimation},
  author={Franzese, Giulio and Bounoua, Mustapha and Michiardi, Pietro},
  booktitle={International Conference on Learning Representations},
  volume={2024},
  pages={16685--16716},
  year={2024}
}

@article{premkumar2025neural,
  title={Neural entropy},
  author={Premkumar, Akhil},
  journal={Advances in Neural Information Processing Systems},
  volume={38},
  pages={21626--21669},
  year={2026}
}

@inproceedings{wang2025informationtheoretic,
  title={Information theoretic text-to-image alignment},
  author={Wang, Chao and Franzese, Giulio and Gallo, Massimo and Michiardi, Pietro and others},
  booktitle={International Conference on Learning Representations},
  volume={2025},
  pages={50574--50610},
  year={2025}
}

@inproceedings{wang2023infodiffusion,
  title={Infodiffusion: Representation learning using information maximizing diffusion models},
  author={Wang, Yingheng and Schiff, Yair and Gokaslan, Aaron and Pan, Weishen and Wang, Fei and De Sa, Christopher and Kuleshov, Volodymyr},
  booktitle={International conference on machine learning},
  pages={36336--36354},
  year={2023},
  organization={PMLR}
}

@article{sclocchi2025phase,
  title={A phase transition in diffusion models reveals the hierarchical nature of data},
  author={Sclocchi, Antonio and Favero, Alessandro and Wyart, Matthieu},
  journal={Proceedings of the National Academy of Sciences},
  volume={122},
  number={1},
  pages={e2408799121},
  year={2025},
  publisher={National Academy of Sciences}
}

@inproceedings{he2023diffusionbert,
  title={Diffusionbert: Improving generative masked language models with diffusion models},
  author={He, Zhengfu and Sun, Tianxiang and Tang, Qiong and Wang, Kuanning and Huang, Xuan-Jing and Qiu, Xipeng},
  booktitle={Proceedings of the 61st annual meeting of the association for computational linguistics (volume 1: Long papers)},
  pages={4521--4534},
  year={2023}
}

@article{kahouli2024molecular,
  title={Molecular relaxation by reverse diffusion with time step prediction},
  author={Kahouli, Khaled and Hessmann, Stefaan Simon Pierre and M{\"u}ller, Klaus-Robert and Nakajima, Shinichi and Gugler, Stefan and Gebauer, Niklas Wolf Andreas},
  journal={Machine Learning: Science and Technology},
  volume={5},
  number={3},
  pages={035038},
  year={2024},
  publisher={IOP Publishing}
}

@article{yang2026stable,
  title={Stable Velocity: A Variance Perspective on Flow Matching},
  author={Yang, Donglin and Zhang, Yongxing and Yu, Xin and Hou, Liang and Tao, Xin and Wan, Pengfei and Qi, Xiaojuan and Liao, Renjie},
  journal={arXiv preprint arXiv:2602.05435},
  year={2026}
}

@inproceedings{
xu2023stable,
title={Stable Target Field for Reduced Variance Score Estimation in Diffusion Models},
author={Yilun Xu and Shangyuan Tong and Tommi S. Jaakkola},
booktitle={The Eleventh International Conference on Learning Representations },
year={2023},
url={https://openreview.net/forum?id=WmIwYTd0YTF}
}

@article{shannon1948mathematical,
  title={A mathematical theory of communication},
  author={Shannon, Claude Elwood},
  journal={The Bell system technical journal},
  volume={27},
  number={3},
  pages={379--423},
  year={1948},
  publisher={Nokia Bell Labs}
}

@article{shi2024simplified,
  title={Simplified and generalized masked diffusion for discrete data},
  author={Shi, Jiaxin and Han, Kehang and Wang, Zhe and Doucet, Arnaud and Titsias, Michalis},
  journal={Advances in neural information processing systems},
  volume={37},
  pages={103131--103167},
  year={2024}
}

@article{sahoo2024simple,
  title={Simple and effective masked diffusion language models},
  author={Sahoo, Subham S and Arriola, Marianne and Schiff, Yair and Gokaslan, Aaron and Marroquin, Edgar and Chiu, Justin T and Rush, Alexander and Kuleshov, Volodymyr},
  journal={Advances in Neural Information Processing Systems},
  volume={37},
  pages={130136--130184},
  year={2024}
}

@article{pillutla2021mauve,
  title={Mauve: Measuring the gap between neural text and human text using divergence frontiers},
  author={Pillutla, Krishna and Swayamdipta, Swabha and Zellers, Rowan and Thickstun, John and Welleck, Sean and Choi, Yejin and Harchaoui, Zaid},
  journal={Advances in Neural Information Processing Systems},
  volume={34},
  pages={4816--4828},
  year={2021}
}

@article{gao2020pile,
  title={The pile: An 800gb dataset of diverse text for language modeling},
  author={Gao, Leo and Biderman, Stella and Black, Sid and Golding, Laurence and Hoppe, Travis and Foster, Charles and Phang, Jason and He, Horace and Thite, Anish and Nabeshima, Noa and others},
  journal={arXiv preprint arXiv:2101.00027},
  year={2020}
}

@article{radford2019language,
  title={Language Models are Unsupervised Multitask Learners},
  author={Radford, Alec and Wu, Jeff and Child, Rewon and Luan, David and Amodei, Dario and Sutskever, Ilya},
  journal={OpenAI Blog},
  year={2019}
}

@article{esteves2026spectrally,
  title={Spectrally-Guided Diffusion Noise Schedules},
  author={Esteves, Carlos and Makadia, Ameesh},
  journal={arXiv preprint arXiv:2603.19222},
  year={2026}
}

@article{dasilva2026designing,
  title={Designing synthetic regulatory elements using the generative AI framework DNA-Diffusion},
  author={DaSilva, Lucas Ferreira and Senan, Simon and Kribelbauer-Swietek, Judith F and Patel, Zain Munir and Louis, Lithin Karmel and Reddy, Aniketh Janardhan and Gabbita, Sameer and Rosen, Jonathan D and Nussbaum, Zach and C{\'o}rdova, C{\'e}sar Miguel Valdez and others},
  journal={Nature Genetics},
  volume={58},
  number={1},
  pages={180--194},
  year={2026},
  publisher={Nature Publishing Group US New York}
}


\appendix

\phantomsection

\section*{Supplementary material}
\label{app:supplementary_overview}

The supplement provides the derivations and implementation details underlying the allocation view, the information-profile target, and the online estimator used in the experiments.

\begin{itemize}[leftmargin=*]
    \item \Cref{app:schedule_unification} rewrites diffusion and flow-matching objectives in a common \(x\)-prediction view, separating the sampling density from the loss weight and deriving the induced allocation \(\phi_x=\pi w_x\) for the schedules used in the paper.

    \item \Cref{app:immse} derives the conditional-entropy-rate profile for affine-Gaussian corruption paths from I--MMSE, and gives the finite-dataset Bayes quantities and two-point decision-window example used to interpret localized uncertainty resolution.

    \item \Cref{app:experimental_details} specifies the online estimator, sampler rebuilds, warm-up, smoothing, endpoint regularization, datasets, architectures, optimization settings, samplers, NFE conventions, guidance settings, metrics, and compute-to-target protocol.

    \item \Cref{app:additional_diagnostics} provides allocation plots, profile galleries, empirical Bayes visualizations, qualitative samples, and consistency checks for the binned loss estimate.
\end{itemize}
 
\section{Noise schedules and weights in a common allocation view}
\label[appendix]{app:schedule_unification}

Diffusion and flow-matching methods are often written in different path coordinates, prediction targets, and loss weights. Directly comparing their schedules therefore confounds how often path locations are sampled with how strongly their errors are weighted. For the affine-Gaussian paths used in our baselines, these objectives can be rewritten in a common \(x\)-prediction view indexed by relative noise or log-SNR. Following the design-space view of \citet{karras2022elucidating} and the weighted log-SNR view of \citet{kingma2023understanding}, this appendix derives the induced allocation
$\phi_x=\pi w_x$
for the fixed schedules used in our experiments.

\subsection{Affine-Gaussian paths}
\label{app:common_path}

For a linear forward SDE, the perturbation kernel can be written as
\begin{equation}
    \rvx_t = s(t)\rvx_0 + s(t)\sigma(t)\bepsilon,
    \qquad
    \bepsilon \sim \mathcal N(\mathbf 0,\I),
\label{eq:app_song_kernel}
\end{equation}
with
\begin{equation}
    s(t)=\exp\!\left(\int_0^t f(\xi)\,d\xi\right),
    \qquad
    \sigma(t)^2=\int_0^t \frac{g(\xi)^2}{s(\xi)^2}\,d\xi .
\label{eq:app_song_s_sigma}
\end{equation}
Here \(s(t)\) is the signal scale and \(\sigma(t)\) is the relative noise scale in the standardized kernel notation of \citet{karras2022elucidating}. For allocation comparisons, the induced kernel is the relevant object. After a monotone reparameterization \(t=t(u)\), we write the same path as
\begin{equation}
    \rvx_u = a(u)\rvx_0 + b(u)\bepsilon,
    \qquad
    a(u)=s(t(u)),
    \qquad
    b(u)=s(t(u))\sigma(t(u)),
    \qquad
    a(u),b(u)>0 .
\label{eq:app_affine_path}
\end{equation}
The coordinate \(u\) indexes path locations. This affine-Gaussian form covers the standard VE, VP, and sub-VP perturbation kernels up to endpoint limits. EDM is included through the VE-style perturbation kernel, and our flow-matching baselines through affine-Gaussian conditional interpolants.

\subsection{Standardized denoising coordinate}
\label{app:canonical_allocation}

Factoring out the signal scale rewrites \cref{eq:app_affine_path} in the Karras-style standardized form
\begin{equation}
    \rvx_u
    =
    a(u)\left(\rvx_0+\frac{b(u)}{a(u)}\bepsilon\right)
    =
    a(u)\bigl(\rvx_0+\sigma(u)\bepsilon\bigr),
\label{eq:app_karras_form}
\end{equation}
with
\begin{equation}
    \sigma(u)\coloneqq \frac{b(u)}{a(u)},
    \qquad
    \gamma(u)\coloneqq \frac{a(u)^2}{b(u)^2}=\frac{1}{\sigma(u)^2},
    \qquad
    \lambda(u)\coloneqq \log\gamma(u)=-2\log\sigma(u).
\label{eq:app_lambda_sigma}
\end{equation}
For allocation comparisons in the denoising coordinate, the deterministic scale \(a(u)\) does not change the posterior inference problem at fixed \(u\).
\begin{equation}
    \bar{\rvx}_u
    \coloneqq
    \frac{\rvx_u}{a(u)}
    =
    \rvx_0+\sigma(u)\bepsilon .
\label{eq:app_karras_standardized}
\end{equation}
Thus the denoising problem can be indexed by relative noise \(\sigma(u)\), or equivalently by log-SNR \(\lambda(u)\).

A schedule is usually specified in an implementation coordinate \(\eta\), not directly in \(\lambda\). If \(\eta\sim\pi_\eta\) and \(\lambda=\lambda(\eta)\) is monotone, the induced density over log-SNR is
\begin{equation}
    \pi(\lambda)
    =
    \pi_\eta(\eta(\lambda))
    \left|
    \frac{d\eta}{d\lambda}
    \right|.
\label{eq:app_pi_lambda}
\end{equation}
The schedule rule samples its native coordinate, while \(\pi(\lambda)\) describes how often training visits denoising problems in the common coordinate.

In the \(x\)-prediction view, a weighted denoising objective has the form
\begin{equation}
    \mathcal L_x
    =
    \int
    \pi(\lambda)w_x(\lambda)
    \,
    \E\!\left[
        \norm{
        \rvx_0-\hat{\rvx}_0(\bar{\rvx}_\lambda,\lambda)
        }_2^2
        \mid \lambda
    \right]
    d\lambda .
\label{eq:app_loss_x}
\end{equation}
The effective allocation is
\begin{equation}
    \phi_x(\lambda)
    \coloneqq
    \pi(\lambda)w_x(\lambda).
\label{eq:app_phi_x}
\end{equation}
The schedule determines how often a denoising problem is visited. The loss weight determines how strongly its error contributes. Their product is the allocation compared across schedules.

\subsection{Continuous-time VDM objective and entropy rate}
\label{app:vdm_entropy_rate}

The continuous-time VDM objective of \citet{kingma2021variational} provides a useful likelihood-based reference point. Let
\[
    \rvz_t
    =
    a(t)\rvx_0+b(t)\bepsilon,
    \qquad
    \gamma(t)=\frac{a(t)^2}{b(t)^2},
    \qquad
    \gamma'(t)\le0 .
\]
The diffusion term of the continuous-time bound can be written in \(x\)-prediction form as
\begin{equation}
    \mathcal L_{\infty}(\theta)
    =
    -\frac12
    \int_0^1
    \gamma'(t)
    \E_{\rvx_0,\bepsilon}
    \left[
    \norm{
    \rvx_0-\hat{\rvx}_{\theta}(\rvz_t;t)
    }_2^2
    \right]
    \,\dd t .
\label{eq:app_vdm_continuous_x}
\end{equation}
The coefficient \(-\frac12\gamma'(t)\) is induced by the continuous-time ELBO and the SNR parameterization of the path. It is not an additional loss weight.

At the Bayes \(x\)-prediction denoiser,
\[
    \hat{\rvx}_{\theta^\star}(\rvz_t;t)
    =
    \E[\rvx_0\mid \rvz_t],
\]
the expected denoising loss equals
\[
    \mmse(t)
    =
    \E
    \left[
    \norm{
    \rvx_0-\E[\rvx_0\mid \rvz_t]
    }_2^2
    \right].
\]
After standardization,
\[
    \frac{\rvz_t}{b(t)}
    =
    \sqrt{\gamma(t)}\,\rvx_0+\bepsilon .
\]
The Gaussian-channel I--MMSE identity gives
\begin{equation}
    \frac{\dd}{\dd t}
    \Hent[\rvx_0\mid \rvz_t]
    =
    -\frac12
    \gamma'(t)\mmse(t).
\label{eq:app_vdm_entropy_rate}
\end{equation}
Thus, at the Bayes denoiser, the continuous-ELBO integrand equals the conditional-entropy rate. For a learned denoiser, the integrand remains model-dependent because the denoising loss need not equal \(\mmse(t)\).

\subsection{Prediction-space conversions}
\label{app:prediction_space_conversions}

Many objectives are written in \(\epsilon\)-prediction form. From \cref{eq:app_karras_standardized},
\begin{equation}
    \bepsilon
    =
    \frac{\bar{\rvx}_u-\rvx_0}{\sigma(u)} .
\label{eq:app_true_eps}
\end{equation}
Given an \(\epsilon\)-prediction model \(\hat{\bepsilon}_\theta(\bar{\rvx}_u;u)\), define its induced \(x\)-reconstruction as
\begin{equation}
    \xhat^{(\epsilon)}(\bar{\rvx}_u;u)
    \coloneqq
    \bar{\rvx}_u
    -
    \sigma(u)\hat{\bepsilon}_\theta(\bar{\rvx}_u;u).
\label{eq:app_eps_induced_x}
\end{equation}
Then
\begin{equation}
    \rvx_0-\xhat^{(\epsilon)}
    =
    \sigma(u)
    \left(
    \hat{\bepsilon}_\theta-\bepsilon
    \right),
\end{equation}
and therefore
\begin{equation}
    \norm{\bepsilon-\hat{\bepsilon}_\theta}_2^2
    =
    \frac{1}{\sigma(u)^2}
    \norm{\rvx_0-\xhat^{(\epsilon)}}_2^2
    =
    e^{\lambda(u)}
    \norm{\rvx_0-\xhat^{(\epsilon)}}_2^2 .
\label{eq:app_eps_to_x}
\end{equation}

An \(\epsilon\)-prediction objective with allocation
\[
    \phi_\epsilon(\lambda)
    =
    \pi(\lambda)w_\epsilon(\lambda)
\]
therefore induces, in the common \(x\)-reconstruction view,
\begin{equation}
    \phi_x(\lambda)
    =
    e^\lambda \phi_\epsilon(\lambda)
    =
    e^\lambda \pi(\lambda)w_\epsilon(\lambda).
\label{eq:app_phi_conversion}
\end{equation} 

\subsection{Common support and plotting coordinate}
\label{app:truncation}

All allocation comparisons use the same finite log-SNR interval
\([\lambda_{\min},\lambda_{\max}]\). If \(\pi(\lambda)\) is the density on its full support, the truncated training density is
\begin{equation}
    \tilde\pi(\lambda)
    =
    \frac{
        \pi(\lambda)\mathbf 1\{\lambda\in[\lambda_{\min},\lambda_{\max}]\}
    }{
        \int_{\lambda_{\min}}^{\lambda_{\max}}\pi(\ell)\,d\ell
    } .
\label{eq:app_truncated_density}
\end{equation}
The corresponding unnormalized effective allocation is
\begin{equation}
    \tilde\phi_x(\lambda)
    =
    \tilde\pi(\lambda)w_x(\lambda),
\label{eq:app_truncated_allocation}
\end{equation}
and allocation plots normalize \(\tilde\phi_x\) on the common support.

For plots shown on a logarithmic \(\sigma\)-axis, using
\(\lambda=-2\log\sigma\), the density with respect to \(\sigma\) is
\begin{equation}
    \psi_x(\sigma)
    =
    \phi_x(\lambda(\sigma))
    \left|
    \frac{d\lambda}{d\sigma}
    \right|
    =
    \frac{2}{\sigma}\phi_x(-2\log\sigma).
\label{eq:app_phi_sigma}
\end{equation}
The density displayed per unit \(\log\sigma\) is proportional to
\(\sigma\psi_x(\sigma)\), equivalently to \(\phi_x(-2\log\sigma)\).

\subsection{Baseline allocations}
\label{app:baseline_allocations}

We instantiate the schedule densities used in the main experiments in the common log-SNR coordinate \(\lambda\). All expressions below are densities with respect to \(\lambda\), shown up to normalization on the common support. When a schedule is used under a fixed \(x\)-prediction objective, its effective allocation is obtained by multiplying the listed \(\pi(\lambda)\) by the experiment's fixed \(x\)-space weight \(w_x(\lambda)\).

\paragraph{EDM.}
EDM~\citep{karras2022elucidating} samples
\begin{equation}
    \log\sigma \sim \mathcal N(P_{\mathrm{mean}},P_{\mathrm{std}}^2).
\end{equation}
Since \(\lambda=-2\log\sigma\), this induces
\begin{equation}
    \pi_{\mathrm{EDM}}(\lambda)
    =
    \mathcal N\!\left(
        \lambda;\,-2P_{\mathrm{mean}},(2P_{\mathrm{std}})^2
    \right).
\label{eq:app_edm_pi_lambda}
\end{equation}
With the EDM \(x\)-space denoising weight
\begin{equation}
    w_x^{\mathrm{EDM}}(\sigma)
    =
    \frac{\sigma^2+\sigma_{\mathrm{data}}^2}
         {\sigma^2\sigma_{\mathrm{data}}^2},
\label{eq:app_edm_weight_sigma}
\end{equation}
we obtain, up to a constant independent of \(\lambda\),
\begin{equation}
    w_x^{\mathrm{EDM}}(\lambda)
    \propto
    1+\sigma_{\mathrm{data}}^2 e^\lambda,
    \qquad
    \phi_x^{\mathrm{EDM}}(\lambda)
    \propto
    \pi_{\mathrm{EDM}}(\lambda)
    \bigl(1+\sigma_{\mathrm{data}}^2 e^\lambda\bigr).
\label{eq:app_edm_phi_lambda}
\end{equation}

\paragraph{Log-uniform in \(\sigma\).}
A log-uniform schedule satisfies
\begin{equation}
    \pi_{\mathrm{LU}}(\sigma)
    =
    \frac{1}{\sigma\log(\sigma_{\max}/\sigma_{\min})}.
\label{eq:app_lu_pi_sigma}
\end{equation}
Since \(\lambda=-2\log\sigma\), the induced density \(\pi_{\mathrm{LU}}(\lambda)\) is uniform on
\[
    [-2\log\sigma_{\max},-2\log\sigma_{\min}].
\]
Therefore, under a fixed \(x\)-prediction objective,
\[
    \phi_x^{\mathrm{LU}}(\lambda)
    =
    \pi_{\mathrm{LU}}(\lambda)w_x(\lambda)
    \propto
    w_x(\lambda).
\]
In our experiments using constant \(x\)-space loss weight, this gives a uniform effective allocation in \(\lambda\). Log-uniform sampling therefore serves as the allocation-neutral reference.
 
\paragraph{Cosine / iDDPM.}
For the continuous cosine limit of \citet{nichol2021improved},
\begin{equation}
    \bar\alpha(t)=\cos^2\!\left(\frac{\pi t}{2}\right),
    \qquad
    \sigma(t)=\tan\!\left(\frac{\pi t}{2}\right),
    \qquad
    t\in(0,1).
\end{equation}
Uniform sampling in \(t\) induces
\begin{equation}
    \pi_{\mathrm{cos}}(\lambda)
    \propto
    \left|
    \frac{dt}{d\lambda}
    \right|
    \propto
    \sech\!\left(\frac{\lambda}{2}\right).
\label{eq:app_cosine_pi_lambda}
\end{equation}
If cosine is used with unit-weight \(\epsilon\)-prediction, \cref{eq:app_phi_conversion} gives
\begin{equation}
    \phi_x^{\mathrm{cos}}(\lambda)
    \propto
    e^\lambda\sech(\lambda/2).
\label{eq:app_cosine_phi_x_lambda}
\end{equation}
If cosine is used only as a schedule under a fixed \(x\)-prediction objective, its allocation is instead
\[
    \phi_x^{\mathrm{cos}}(\lambda)
    \propto
    \pi_{\mathrm{cos}}(\lambda)w_x(\lambda).
\]

\paragraph{Flow matching with the OT path.}
For the OT interpolation~\citep{lipman2023flow},
\begin{equation}
    \rvx_t=(1-t)\rvx_0+t\bepsilon,
    \qquad
    t\in(0,1),
\end{equation}
the relative-noise and log-SNR coordinates are
\begin{equation}
    \sigma(t)=\frac{t}{1-t},
    \qquad
    \lambda(t)=\log\sigma(t)^{-2}
    =
    2\log\frac{1-t}{t}.
\end{equation}
Uniform sampling in \(t\) induces
\begin{equation}
    \pi_{\mathrm{OT}}(\lambda)
    \propto
    \left|
    \frac{dt}{d\lambda}
    \right|
    \propto
    \sech^2(\lambda/4).
\label{eq:app_ot_pi_lambda}
\end{equation}
For the velocity target \(v=\bepsilon-\rvx_0\),
\begin{equation}
    \rvx_0=\rvx_t-tv,
    \qquad
    \hat{\rvx}_0=\rvx_t-t\hat v .
\end{equation}
Thus
\begin{equation}
    \norm{v-\hat v}_2^2
    =
    \frac{1}{t^2}
    \norm{\rvx_0-\hat{\rvx}_0}_2^2 .
\label{eq:app_fmot_v_to_x}
\end{equation}
The unit-weight velocity objective therefore induces
\begin{equation}
    w_x^{\mathrm{FM\text{-}OT}}(\lambda)
    =
    \frac{1}{t(\lambda)^2}
    =
    \left(1+e^{\lambda/2}\right)^2 .
\label{eq:app_ot_weight_lambda}
\end{equation}
Combining this weight with \(\pi_{\mathrm{OT}}(\lambda)\), and using
\[
\sech^2(\lambda/4)
\propto
\frac{e^{\lambda/2}}{(1+e^{\lambda/2})^2},
\]
gives
\begin{equation}
    \phi_x^{\mathrm{FM\text{-}OT}}(\lambda)
    =
    \pi_{\mathrm{OT}}(\lambda)
    w_x^{\mathrm{FM\text{-}OT}}(\lambda)
    \propto
    e^{\lambda/2}.
\label{eq:app_ot_phi_lambda}
\end{equation}

\begin{figure}
    \centering
    \includegraphics[width=\linewidth]{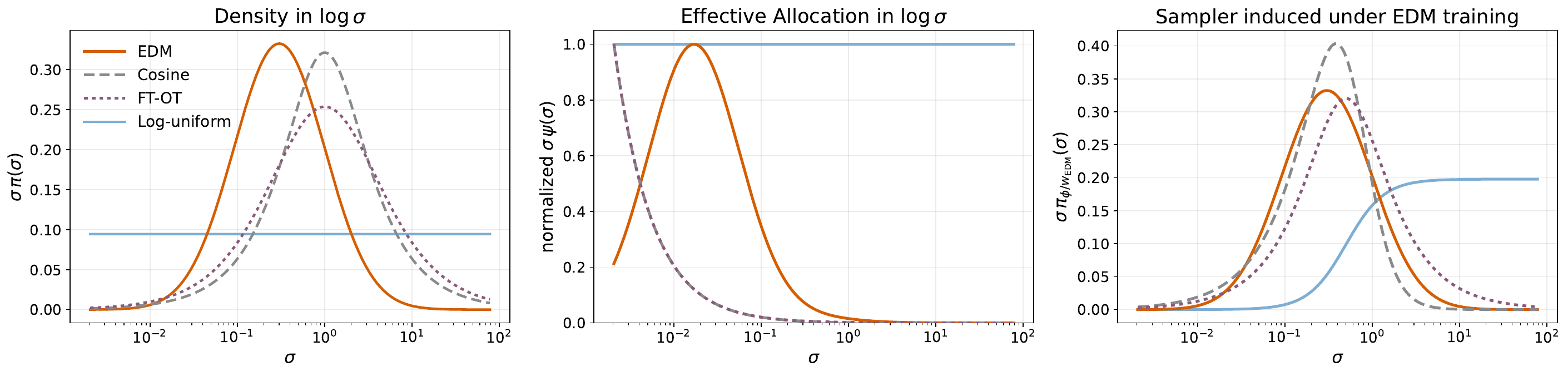}
    \caption{
    \textbf{Baseline schedules induce different effective allocations.}
    Curves are expressed in the common \(x\)-prediction view on the shared log-SNR support. The plotted allocation combines the sampling density and the \(x\)-space loss weight, \(\phi_x(\lambda)=\pi(\lambda)w_x(\lambda)\).
    }
    \label{fig:app_schedule_allocations}
\end{figure}

\begin{table}[t]
\centering
\caption{
\textbf{Baseline allocations in the common \(x\)-prediction view.}
Densities are with respect to \(\lambda=\log\mathrm{SNR}\) and are shown up to normalization on the common support. Symbolic \(w_x(\lambda)\) denotes the fixed \(x\)-space loss weight of the corresponding experiment.
}
\label{tab:app_schedule_summary}
\small
\setlength{\tabcolsep}{5pt}
\renewcommand{\arraystretch}{1.15}
\resizebox{\columnwidth}{!}{%
\begin{tabular}{llll}
\toprule
Baseline & Sampling rule & \(\pi(\lambda)\) & \(\phi_x(\lambda)\) \\
\midrule
EDM
&
\(\log\sigma \sim \mathcal N(P_{\rm mean},P_{\rm std}^2)\)
&
\(\mathcal N(\lambda;-2P_{\rm mean},(2P_{\rm std})^2)\)
&
\(\propto \pi_{\rm EDM}(\lambda)(1+\sigma_{\rm data}^2 e^\lambda)\)
\\

Log-uniform
&
\(\log\sigma\sim\mathcal U\)
&
const.
&
\(\propto w_x(\lambda)\)
\\

Cosine / iDDPM
&
\(t\sim\mathcal U(0,1)\)
&
\(\propto \sech(\lambda/2)\)
&
\(\propto \pi_{\rm cos}(\lambda)w_x(\lambda)\)
\\

FM-OT
&
\(t\sim\mathcal U(0,1)\)
&
\(\propto \sech^2(\lambda/4)\)
&
\(\propto e^{\lambda/2}\)
\\
\bottomrule
\end{tabular}%
}
\end{table} 
\section{Conditional-entropy rates and I--MMSE}
\label{app:immse}

The information profile used by \method{} is the conditional-entropy rate along an affine-Gaussian corruption path. This appendix derives the rate from I--MMSE, gives the VE and log-SNR forms used in the paper, and records finite-dataset Bayes formulas together with the two-point decision-window example.

\subsection{Affine-Gaussian paths and entropy rate}
\label{app:immse_core}

Consider the affine-Gaussian corruption path
\begin{equation}
    \rvx_u
    =
    a(u)\rvx_0+b(u)\bepsilon,
    \qquad
    \rvx_0\sim p_{\mathrm{data}},
    \qquad
    \bepsilon\sim\mathcal N(\mathbf 0,\I_d),
    \qquad
    a(u),b(u)>0 .
\label{eq:app_affine_gaussian_channel}
\end{equation}
For fixed \(u\), dividing by the noise scale is an invertible transformation of the observation,
\begin{equation}
    \tilde{\rvx}_u
    \coloneqq
    \frac{\rvx_u}{b(u)}
    =
    \sqrt{\gamma(u)}\,\rvx_0+\bepsilon,
    \qquad
    \gamma(u)\coloneqq \frac{a(u)^2}{b(u)^2}.
\label{eq:app_standardized_affine_channel}
\end{equation}
Thus \(\rvx_u\) and \(\tilde{\rvx}_u\) define the same posterior over \(\rvx_0\), and
\begin{equation}
    \Hent[\rvx_0\mid \rvx_u]
    =
    \Hent[\rvx_0\mid \tilde{\rvx}_u].
\label{eq:app_entropy_invariance_affine}
\end{equation}

Define the Bayes denoising error at SNR \(\gamma\) by
\begin{equation}
    \mmse(\gamma)
    \coloneqq
    \E\!\left[
        \norm{
        \rvx_0-\E[\rvx_0\mid \sqrt{\gamma}\rvx_0+\bepsilon]
        }_2^2
    \right].
\label{eq:app_mmse_gamma_def}
\end{equation}
The Gaussian-channel I--MMSE identity~\citep{guo2005mutual,palomar2005gradient} gives
\begin{equation}
    \frac{\partial}{\partial \gamma}
    I(\rvx_0;\sqrt{\gamma}\rvx_0+\bepsilon)
    =
    \frac12\,\mmse(\gamma).
\label{eq:app_immse_gamma}
\end{equation}
Since
\[
    \Hent[\rvx_0\mid \rvx_u]
    =
    \Hent[\rvx_0]
    -
    I(\rvx_0;\tilde{\rvx}_u),
\]
the chain rule yields
\begin{equation}
    \frac{d}{du}
    \Hent[\rvx_0\mid \rvx_u]
    =
    -\frac12\,\gamma'(u)\,\mmse(\gamma(u)).
\label{eq:app_general_entropy_rate}
\end{equation}
The conditional-entropy rate is therefore determined by the Bayes-optimal denoising error at the current SNR and by how quickly the path changes SNR.

For an arbitrary monotone coordinate, we define the pathwise information profile as the magnitude of this derivative,
\begin{equation}
    \rho^\star_u(u)
    \propto
    \left|
    \frac{d}{du}
    \Hent[\rvx_0\mid \rvx_u]
    \right|
    =
    \frac12
    \left|
    \gamma'(u)
    \right|
    \mmse(\gamma(u)).
\label{eq:app_entropy_rate_profile}
\end{equation}
Large values mark path locations where posterior uncertainty about \(\rvx_0\) changes most rapidly in the chosen coordinate.

\subsection{VE and log-SNR forms}
\label{app:ve_entropy_rate}

For the VE channel,
\begin{equation}
    \rvx_\sigma
    =
    \rvx_0+\sigma\bepsilon,
    \qquad
    \bepsilon\sim\mathcal N(\mathbf 0,\I_d),
\label{eq:app_ve_channel}
\end{equation}
we have \(a(\sigma)\equiv 1\), \(b(\sigma)=\sigma\), and \(\gamma(\sigma)=\sigma^{-2}\). Substituting into \cref{eq:app_general_entropy_rate} gives
\begin{equation}
    \frac{d}{d\sigma}
    \Hent[\rvx_0\mid \rvx_\sigma]
    =
    \frac{\mmse(\sigma)}{\sigma^3},
\label{eq:app_entropy_rate_sigma}
\end{equation}
where
\begin{equation}
    \mmse(\sigma)
    \coloneqq
    \E\!\left[
        \norm{
        \rvx_0-\E[\rvx_0\mid \rvx_\sigma]
        }_2^2
    \right].
\label{eq:app_mmse_sigma_def}
\end{equation}
Equivalently, with log-SNR \(\lambda=\log\gamma=-2\log\sigma\),
\begin{equation}
    -\frac{d}{d\lambda}
    \Hent[\rvx_0\mid \rvx_\sigma]
    =
    \frac12\,\gamma\,\mmse(\gamma)
    =
    \frac12\,\frac{\mmse(\sigma)}{\sigma^2}.
\label{eq:app_entropy_rate_lambda}
\end{equation}
When training is parameterized by VE noise scale \(\sigma\), \cref{eq:app_entropy_rate_sigma} is the profile estimated online by \method{}.

The profiles in \cref{eq:app_entropy_rate_sigma,eq:app_entropy_rate_lambda} are densities with respect to different path coordinates. Under the change of variables \(\lambda=-2\log\sigma\), the density transforms by the Jacobian \(|d\sigma/d\lambda|=\sigma/2\). Alignment with a training allocation is meaningful only after both quantities are expressed in the same coordinate.

\subsection{Common path examples}
\label{app:immse_path_examples}

The identity in \cref{eq:app_general_entropy_rate} applies to any affine-Gaussian path.

\paragraph{VE.}
For \(a(u)\equiv 1\), \(b(u)=\sigma(u)\),
\begin{equation}
    \gamma(u)=\sigma(u)^{-2},
    \qquad
    \frac{d}{du}
    \Hent[\rvx_0\mid \rvx_u]
    =
    \frac{\sigma'(u)}{\sigma(u)^3}\,
    \mmse(\gamma(u)).
\label{eq:app_ve_general_u}
\end{equation}

\paragraph{VP.}
For
\[
    \rvx_u=\alpha(u)\rvx_0+\sqrt{1-\alpha(u)^2}\,\bepsilon,
\]
we have
\[
    \gamma(u)=\frac{\alpha(u)^2}{1-\alpha(u)^2},
\]
and therefore
\begin{equation}
    \frac{d}{du}
    \Hent[\rvx_0\mid \rvx_u]
    =
    -\frac{\alpha(u)\alpha'(u)}
    {(1-\alpha(u)^2)^2}
    \mmse(\gamma(u)).
\label{eq:app_vp_entropy_rate}
\end{equation}

\paragraph{Affine-Gaussian flow matching.}
For the linear Gaussian interpolant
\[
    \rvx_u=(1-u)\rvx_0+u\bepsilon,
\]
we have
\[
    a(u)=1-u,
    \qquad
    b(u)=u,
    \qquad
    \gamma(u)=\frac{(1-u)^2}{u^2}.
\]
Thus
\begin{equation}
    \frac{d}{du}
    \Hent[\rvx_0\mid \rvx_u]
    =
    \frac{1-u}{u^3}\,
    \mmse(\gamma(u)).
\label{eq:app_fm_entropy_rate}
\end{equation}

\subsection{Empirical Bayes formulas for finite datasets}
\label{app:empirical_bayes}

Under an empirical prior, the Bayes denoiser and posterior covariance have closed-form finite-sum expressions. Given \(\{x_i\}_{i=1}^N\subset\mathbb R^d\), let
\begin{equation}
    p_{\mathrm{data}}(x_0)
    =
    \frac1N
    \sum_{i=1}^N
    \delta(x_0-x_i).
\label{eq:app_empirical_prior}
\end{equation}
Under the VE channel, the noisy marginal is the Gaussian mixture
\begin{equation}
    p(x;\sigma)
    =
    \frac1N
    \sum_{i=1}^N
    \mathcal N(x;x_i,\sigma^2\I_d).
\label{eq:app_empirical_mixture}
\end{equation}
Bayes' rule gives posterior weights
\begin{equation}
    w_i(x;\sigma)
    =
    \frac{
        \mathcal N(x;x_i,\sigma^2\I_d)
    }{
        \sum_{j=1}^N
        \mathcal N(x;x_j,\sigma^2\I_d)
    },
    \qquad
    \sum_{i=1}^N w_i(x;\sigma)=1,
\label{eq:app_empirical_weights}
\end{equation}
and the Bayes denoiser is
\begin{equation}
    \hat{x}^{\star}(x;\sigma)
    =
    \E[\rvx_0\mid \rvx_\sigma=x]
    =
    \sum_{i=1}^N w_i(x;\sigma)x_i.
\label{eq:app_empirical_bayes_mean}
\end{equation}
The posterior covariance is
\begin{equation}
    \Cov(\rvx_0\mid \rvx_\sigma=x)
    =
    \sum_{i=1}^N
    w_i(x;\sigma)
    \bigl(x_i-\hat{x}^{\star}(x;\sigma)\bigr)
    \bigl(x_i-\hat{x}^{\star}(x;\sigma)\bigr)^\top ,
\label{eq:app_empirical_cov}
\end{equation}
so
\begin{equation}
    \mmse(\sigma)
    =
    \E_{x\sim p(\cdot;\sigma)}
    \left[
        \tr\Cov(\rvx_0\mid \rvx_\sigma=x)
    \right].
\label{eq:app_empirical_mmse}
\end{equation}
Substituting this quantity into \cref{eq:app_entropy_rate_sigma} gives the empirical Bayes entropy-rate profile used for offline analysis. These profiles are diagnostics only and are not provided to \method{} during training.

The same empirical posterior gives the score through Tweedie's formula,
\begin{equation}
    \nabla_x\log p(x;\sigma)
    =
    \frac{\hat{x}^{\star}(x;\sigma)-x}{\sigma^2}.
\label{eq:app_score_from_denoiser}
\end{equation}
Stationary points of the noisy marginal therefore satisfy
\begin{equation}
    \nabla_x\log p(x;\sigma)=0
    \quad\Longleftrightarrow\quad
    x=\hat{x}^{\star}(x;\sigma).
\label{eq:app_fixed_point_equation}
\end{equation}

\subsection{Two-point decision window}
\label{app:toy_symmetry}

The information-window behavior can be seen analytically in a symmetric two-point distribution. Let
\begin{equation}
    p_{\mathrm{data}}(x_0)
    =
    \frac12\delta(x_0-a)
    +
    \frac12\delta(x_0+a),
    \qquad
    a>0.
\label{eq:toy_two_point_prior}
\end{equation}
The noisy marginal under the VE channel is
\begin{equation}
    p(x;\sigma)
    =
    \frac12\mathcal N(x;a,\sigma^2)
    +
    \frac12\mathcal N(x;-a,\sigma^2).
\label{eq:toy_marginal}
\end{equation}
The posterior weight on \(+a\) is
\begin{equation}
    w_+(x;\sigma)
    =
    \frac{1}
    {1+\exp\!\left(-\frac{2ax}{\sigma^2}\right)},
    \qquad
    w_-(x;\sigma)=1-w_+(x;\sigma),
\label{eq:toy_weights}
\end{equation}
so the Bayes denoiser is
\begin{equation}
    \hat{x}^{\star}(x;\sigma)
    =
    a\bigl(w_+(x;\sigma)-w_-(x;\sigma)\bigr)
    =
    a\tanh\!\left(\frac{ax}{\sigma^2}\right).
\label{eq:toy_denoiser}
\end{equation}
Since \(x_0\in\{\pm a\}\), the posterior variance is
\begin{equation}
    \Var(x_0\mid x_\sigma=x)
    =
    a^2-\hat{x}^{\star}(x;\sigma)^2
    =
    a^2\sech^2\!\left(\frac{ax}{\sigma^2}\right),
\label{eq:toy_post_var}
\end{equation}
and
\begin{equation}
    \mmse(\sigma)
    =
    \E_{x\sim p(\cdot;\sigma)}
    \left[
        a^2\sech^2\!\left(\frac{ax}{\sigma^2}\right)
    \right].
\label{eq:toy_mmse}
\end{equation}
The VE entropy-rate profile is therefore \(\mmse(\sigma)/\sigma^3\).

Using \cref{eq:app_score_from_denoiser}, the score is
\begin{equation}
    \frac{\partial}{\partial x}\log p(x;\sigma)
    =
    \frac{a\tanh\!\left(\frac{ax}{\sigma^2}\right)-x}{\sigma^2}.
\label{eq:toy_score}
\end{equation}

Stationary points satisfy
\begin{equation}
    x
    =
    a\tanh\!\left(\frac{ax}{\sigma^2}\right).
\label{eq:toy_self_consistency}
\end{equation}
Nonzero fixed-point branches appear when the slope of the right-hand side at the origin exceeds one,
\begin{equation}
    \frac{a^2}{\sigma^2}>1
    \qquad
    \Longleftrightarrow
    \qquad
    \sigma<\sigma_c\coloneqq a .
\label{eq:toy_branch_condition}
\end{equation}
\begin{wrapfigure}{r}{0.46\linewidth}
  \vspace{-1.2em}
  \centering
  \includegraphics[width=\linewidth]{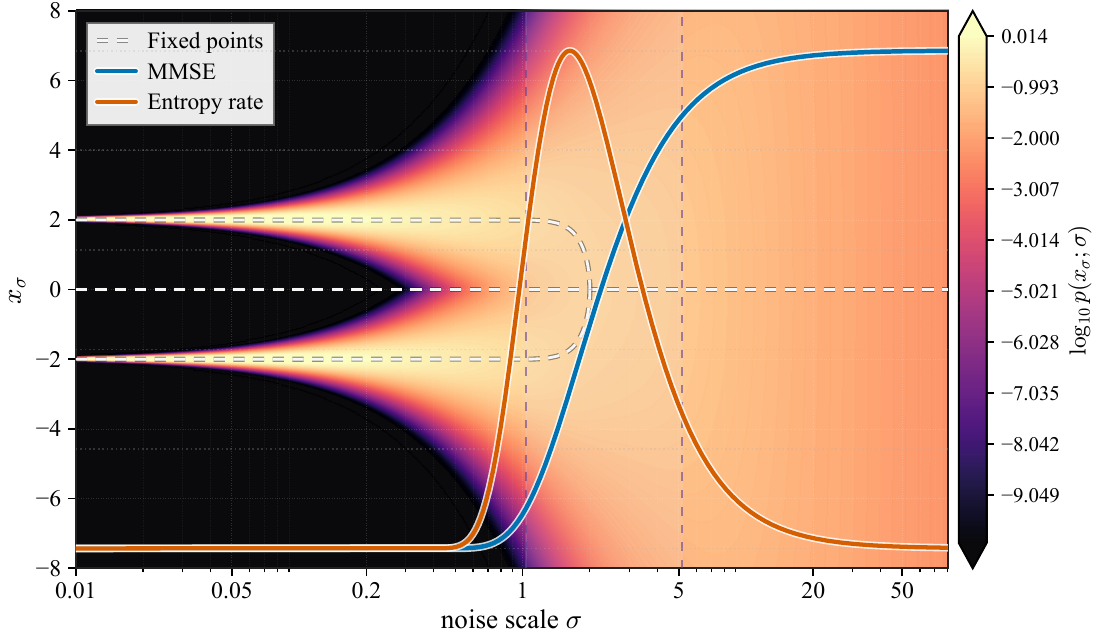}
\caption{
\textbf{Entropy rate localizes the decision window.}
MMSE measures remaining uncertainty; entropy rate marks where that uncertainty changes fastest. Fixed-point branches show the same transition geometrically.
}
  \label{fig:toy_symmetry_window}
  \vspace{-1.0em}
\end{wrapfigure}
Equivalently,
\begin{equation}
    \frac{\partial^2}{\partial x^2}\log p(0;\sigma)
    =
    \frac{a^2-\sigma^2}{\sigma^4},
\label{eq:toy_hessian_origin}
\end{equation}
so \(x=0\) changes from a local maximum of the noisy marginal to a local minimum as \(\sigma\) crosses \(\sigma_c\). The noisy marginal changes from unimodal to bimodal, and the Bayes denoiser stops averaging over both modes and begins committing to one branch. This transition is the analytic version of the information window in the main text. At high noise, observations carry little information about which atom generated the sample. At very low noise, the posterior is already concentrated. The entropy-rate profile peaks in the intermediate region where posterior uncertainty collapses most rapidly.

\begin{figure}[t]
    \centering
    \includegraphics[width=\linewidth]{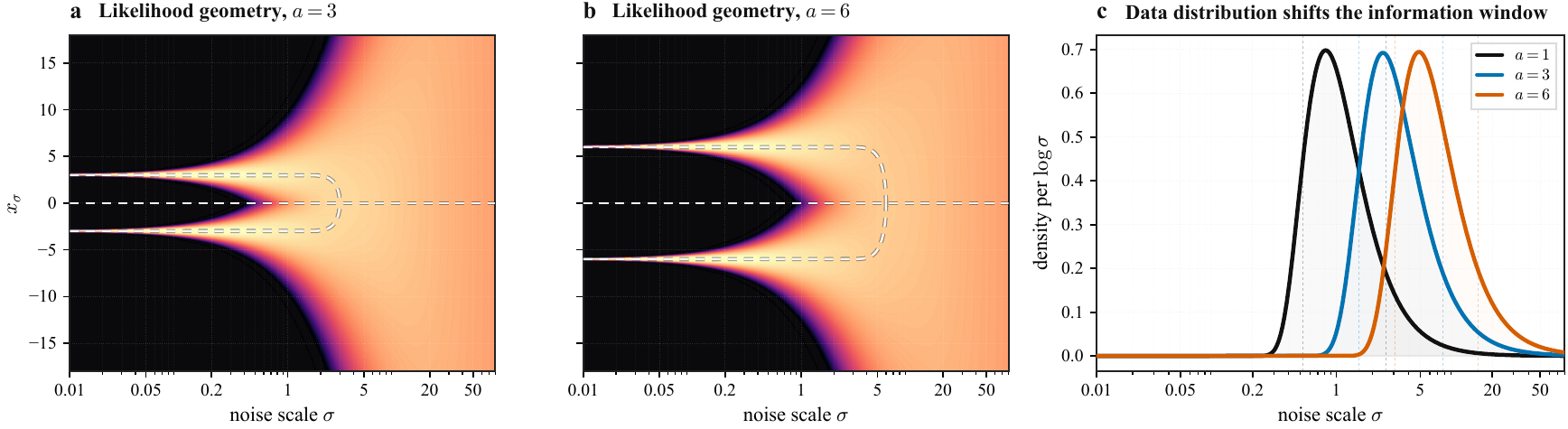}
\caption{
\textbf{The data distribution sets the information window.}
Under the same VE path, changing the atom scale \(a\) in \(x_0\in\{-a,+a\}\) shifts where posterior uncertainty is resolved.
\textbf{(a,b)} Likelihood geometry for \(a=1\) and \(a=6\). Bayes-denoiser fixed points move along the noise axis as data scale changes.
\textbf{(c)} The entropy-rate density shifts with \(a\), showing that the informative region is induced by the data distribution under the corruption path, not the noise coordinate alone.
}
    \label{fig:toy_data_scale_window}
\end{figure}

\paragraph{Data dependence of the decision window.} 
The two-point example also shows that the information window is set by the data distribution under the corruption path, not by the noise coordinate alone. In the prior \(x_0\in\{-a,+a\}\), the only data scale is the atom scale \(a\). Writing \(x=ay\) and \(\sigma=ar\), the posterior weight becomes
\[
    w_+(ay;ar)
    =
    \frac{1}
    {1+\exp\!\left(-\frac{2y}{r^2}\right)} ,
\]
so posterior uncertainty depends on the relative noise level \(r=\sigma/a\). Consequently, the Bayes denoising error has the scaling form
\[
    \mmse_a(\sigma)
    =
    a^2\,m\!\left(\frac{\sigma}{a}\right)
\]
for a dimensionless function \(m\). The VE entropy-rate density is
\[
    \rho_a(\sigma)
    \propto
    \frac{\mmse_a(\sigma)}{\sigma^3}
    =
    \frac{1}{a}
    \frac{
        m(\sigma/a)
    }{
        (\sigma/a)^3
    } .
\]
Thus changing the data scale shifts the information window along the same VE path. \Cref{fig:toy_data_scale_window} shows this shift explicitly.

After normalization as a density over \(\sigma\), the profile has the same shape in the relative coordinate \(\sigma/a\), so its peak shifts linearly with the data scale \(a\). The fixed-point transition occurs at \(\sigma_c=a\), while the entropy-rate maximum occurs at a constant multiple of \(a\). Thus the schedule coordinate fixes how noise is indexed, but the data distribution determines where uncertainty is resolved.
\subsection{Regularization of the estimated information profile}
\label{app:regularization}

\method{} constructs its training sampler from an online estimate of the information profile. Before normalization, this estimate can contain low-noise boundary structure that is not the interior uncertainty-resolution region we want to allocate training around. If left untreated, normalization can place excessive mass near the smallest noise levels and reduce coverage of the region where posterior uncertainty changes. We therefore apply a smooth low-noise gate to the estimated profile before converting it into a sampling density. The gate regularizes the estimate; it is not an additional schedule family.

Let \(r(\sigma)\ge0\) denote the binned and smoothed entropy-rate estimate at a sampler rebuild. The low-noise boundary has different origins for continuous and discrete endpoints. For continuous data, \(\Hent[\rvx_0\mid \rvx_\sigma]\) is a differential entropy. As \(\sigma\to0\), the posterior concentrates with covariance of order \(\sigma^2\), so
\[
    \Hent[\rvx_0\mid \rvx_\sigma]
    =
    d\log\sigma + O(1).
\]
Under standard regularity conditions, \(\mmse(\sigma)=\Theta(\sigma^2)\). The VE entropy-rate identity gives
\begin{equation}
    \frac{\dd}{\dd\sigma}
    \Hent[\rvx_0\mid \rvx_\sigma]
    =
    \frac{\mmse(\sigma)}{\sigma^3}
    =
    \Theta(\sigma^{-1})
    \qquad (\sigma\to0).
\label{eq:app_rate_tail}
\end{equation}
This contribution is finite on the truncated training interval \([\sigma_{\min},\sigma_{\max}]\) with \(\sigma_{\min}>0\), but it can dominate normalization near the lower endpoint. For discrete endpoints, conditional entropy remains bounded as \(\sigma\to0\); empirically, the estimated profile can instead contain an extended low-noise power-law segment,
\begin{equation}
    r(\sigma)\propto \sigma^\alpha,
    \qquad
    \log r(\sigma)=\mathrm{const}+\alpha\log\sigma .
\label{eq:app_discrete_power_law}
\end{equation}
The mechanisms differ, but the practical issue is the same: endpoint structure in the raw estimate can pull allocation away from the interior information-bearing region.

We suppress this boundary structure with the multiplicative gate
\begin{equation}
    g_{c,n}(\sigma)
    =
    \frac{\sigma^n}{\sigma^n+c^n},
    \qquad
    c>0,\quad n\ge 2 .
\label{eq:app_gate_def}
\end{equation}
For \(\sigma\gg c\), \(g_{c,n}(\sigma)\approx1\); for \(\sigma\ll c\), \(g_{c,n}(\sigma)\approx(\sigma/c)^n\). Combining the gate with the continuous low-noise scaling in \cref{eq:app_rate_tail} gives
\begin{equation}
    \left(
    \frac{\dd}{\dd\sigma}
    \Hent[\rvx_0\mid \rvx_\sigma]
    \right)
    g_{c,n}(\sigma)
    =
    \Theta(\sigma^{n-1})
    \qquad (\sigma\to0),
\label{eq:app_gated_tail}
\end{equation}
so the gated profile vanishes at the boundary for \(n\ge2\). For discrete endpoints, the same gate attenuates the empirical low-noise power-law segment. We use \(n=3\) in all experiments.

The regularized allocation target is
\begin{equation}
    \rho(\sigma)
    =
    \frac{
        r(\sigma)g_{c,n}(\sigma)
    }{
        \int_{\sigma_{\min}}^{\sigma_{\max}}
        r(\tilde\sigma)g_{c,n}(\tilde\sigma)\,d\tilde\sigma
    } .
\label{eq:app_gated_profile}
\end{equation}
The sampler is then built from this target using the fixed training weight, as in \cref{eq:pi_from_rho_u},
\[
    \pi(\sigma)
    \propto
    \rho(\sigma)/w(\sigma),
\]
followed by normalization and inverse-CDF sampling. When \(w\) is constant, the sampler density is proportional to \(\rho\).

\paragraph{Gate pivot.}
The gate form and exponent are fixed across experiments, with \(n=3\). The pivot \(c\) is a boundary regularization parameter, not a schedule parameter selected by validation performance. For continuous image experiments, \(c\) is computed from the current profile using an onset-of-information rule. Let \(r(\sigma)\) denote the ungated online estimate and define
\[
    \bar r(\sigma)
    =
    \frac{r(\sigma)}
    {\max_{\tilde\sigma} r(\tilde\sigma)} .
\]
Scanning from high to low noise, we set \(c\) at the first persistent crossing of a fixed threshold \(p\),
\[
    c
    =
    \sup
    \Bigl\{
    \sigma
    \,\Big|\,
    \bar r(\sigma') < p
    \ \text{for all } \sigma'\ge\sigma
    \Bigr\}.
\]
We use \(p=0.002\). On CIFAR-10 this gives \(c\approx0.153\), rounded to \(c=0.15\).

For discrete shifted-regime experiments, the low-noise boundary appears as an extended power-law segment in log--log coordinates. In the reported DNA experiments we use the fixed gate pivot \(c=0.2\), with the same exponent \(n=3\). This pivot is motivated by the observed boundary structure and is kept fixed across runs. The gate regularizes the estimated profile before normalization; it is not an additional schedule family or a validation-tuned sampling distribution.

\begin{figure}[t]
    \centering
    \includegraphics[width=\linewidth]{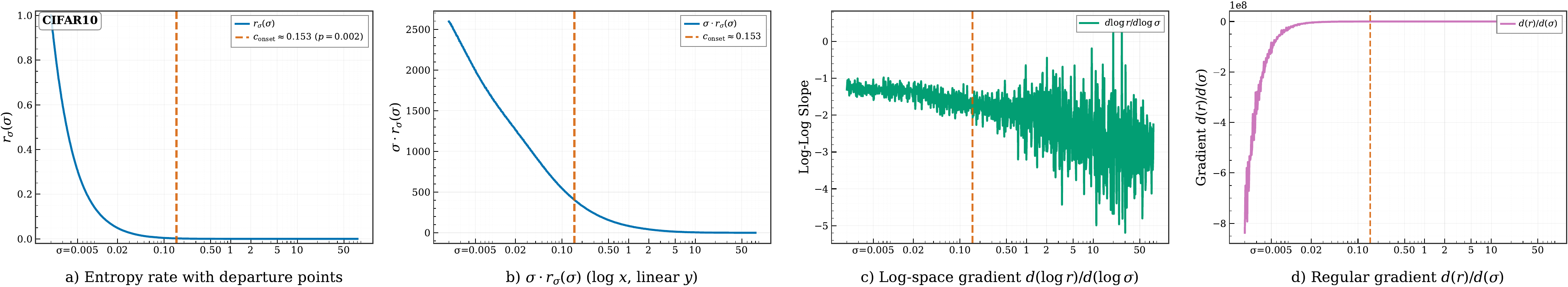}
    \includegraphics[width=\linewidth]{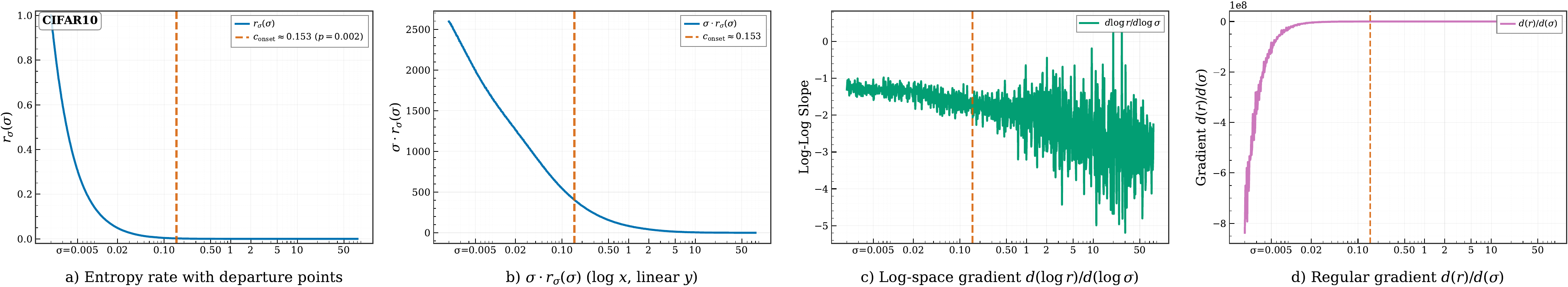}
    \includegraphics[width=\linewidth]{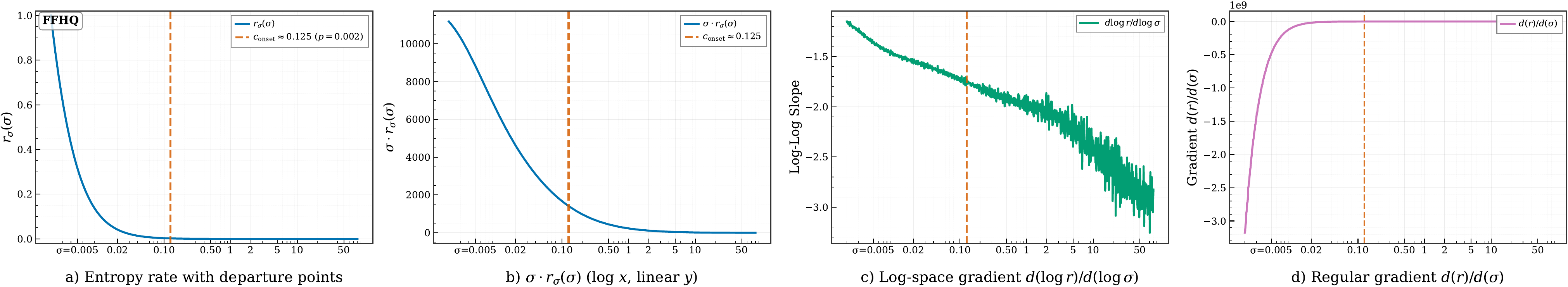}
    \includegraphics[width=\linewidth]{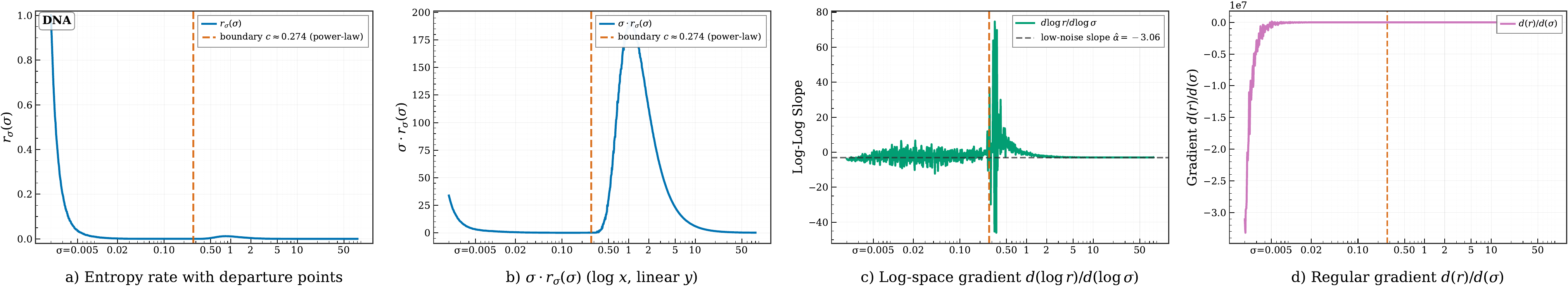}
    \caption{
    \textbf{Low-noise boundary structure motivates gated entropy-rate profiles.}
    Continuous domains show low-noise entropy-rate growth, while discrete domains show extended low-noise power-law structure in the raw estimate. Vertical markers indicate the detected pivot \(c\), the upper edge of the boundary-dominated region. The gate suppresses endpoint structure before normalization while preserving the interior information-bearing region.
    }
    \label{fig:gating_boundary_structure}
\end{figure}
\section{Experimental and reproducibility details}
\label{app:experimental_details}

The experiments isolate training allocation. Within each controlled comparison, the architecture, denoising objective, loss weighting, optimizer, EMA, sampler, inference budget, and evaluation protocol are fixed; only the training noise distribution changes. Image experiments follow the EDM setup of \citet{karras2022elucidating}. Discrete endpoint experiments keep binary or tokenized data at the clean endpoint, but train with a continuous VE Gaussian corruption process.

The main text defines the comparison protocol, compute-to-target metric, and use of offline diagnostic profiles. This appendix provides dataset-specific architectures, training settings, samplers, metrics, and compute resources for reproducibility.

\subsection{\method{} sampler implementation}
\label{app:infonoise_sampler_details}

The online estimator is maintained on a fixed grid in the training coordinate. Each batch records the sampled noise level and the unweighted \(x\)-space denoising error \(\ell\). Losses are binned in \(\log\sigma\), smoothed across refreshes, multiplied by the VE path factor, gated near the low-noise endpoint using \cref{app:regularization}, and normalized to obtain the target profile \(\hat\rho(\sigma)\). The sampler is rebuilt from \(\pi(\sigma)\propto\hat\rho(\sigma)/w(\sigma)\), followed by normalization and inverse-CDF sampling. This keeps the loss weight fixed while adapting the sampling density so that the induced allocation follows the estimated profile.

Before the first reliable estimate is available, \method{} samples from a fixed warm-up prior \(\pi_0\). After the first refresh, all subsequent sampler rebuilds use the online profile estimate. Warm-up length, smoothing, minimum bin counts, refresh cadence, and endpoint gating are fixed within each domain.

\subsection{Image experiments}
\label{app:edm_image_details}

For CIFAR-10 and FFHQ \(64{\times}64\), we use the EDM image setup of \citet{karras2022elucidating} without architecture or optimizer retuning. Images are represented in \([-1,1]\). The denoiser uses the EDM preconditioning coefficients \(c_{\mathrm{in}},c_{\mathrm{skip}},c_{\mathrm{out}},c_{\mathrm{noise}}\) with \(\sigma_{\mathrm{data}}=0.5\), \(\sigma_{\min}=0.002\), and \(\sigma_{\max}=80\). The EDM baseline samples \(\log\sigma\) from the original log-normal training distribution. \method{} keeps the denoiser, preconditioning, loss weight, optimizer, EMA, augmentations, and sampler fixed, and changes only the training distribution over \(\sigma\) after warm-up.

Sampling uses the deterministic EDM probability-flow sampler with a second-order Heun solver and the \(\rho=7\) noise grid. All schedules use the same solver, noise range, checkpoint budget, and FID protocol. We use NFE \(=35\) for CIFAR-10 and NFE \(=79\) for FFHQ \(64{\times}64\). FID is computed from \(50{,}000\) generated samples against the standard real-image statistics.

\subsection{Discrete endpoint experiments}
\label{app:discrete_sequence_details}

The shifted representation and sequence experiments keep the clean endpoint discrete while using a continuous VE Gaussian corruption process. For binary endpoints \(\rvx_0\in\{0,1\}^D\), the network outputs Bernoulli logits at each position, which are converted to probabilities
\[
    \hat{\rvx}_0(\rvx_\sigma;\sigma)
    =
    \operatorname{sigmoid}(\ell_\theta(\rvx_\sigma,\sigma)).
\]
Training uses probability-space denoising with the same fixed loss weight within each schedule comparison. For sampling, the probability denoiser gives the VE score estimate
\[
    \widehat{\nabla_{\rvx_\sigma}\log p(\rvx_\sigma;\sigma)}
    =
    \frac{\hat{\rvx}_0(\rvx_\sigma;\sigma)-\rvx_\sigma}{\sigma^2},
\]
which is integrated with the same second-order Heun solver across schedules. Samples are decoded by thresholding the final probabilities at \(0.5\).

Within each dataset, schedule variants use the same architecture, optimizer, objective, sampler, and evaluation protocol. EDM transfer uses the image-domain EDM log-normal sampler with \(P_{\mathrm{mean}}=-1.2\), \(P_{\mathrm{std}}=1.2\), and \(w_{\mathrm{EDM}}\). Log-uniform sampling uses \(w(\sigma)\equiv1\). \method{} also uses \(w(\sigma)\equiv1\), but replaces the fixed sampler after warm-up with the online information-profile estimate. Self-conditioning and classifier-free guidance are used only in the text experiments.

\paragraph{Binarized CIFAR-10.}
CIFAR-10 images are converted to flattened Gray-coded RGB bitstreams. We use \(45{,}000\) training images, \(5{,}000\) validation images, and the standard test split. Each \(32\times32\times3\) image is represented with \(24\) bits per pixel, giving \(D=24{,}576\). Bits are flattened in raster order. The model is a multiscale 1D U-Net transformer with base width \(64\), time embedding dimension \(128\), head dimension \(32\), and \(8\) levels. The hierarchy uses downsampling factors \((2,2,2,2,2,2,2)\), width multipliers \((1,1,2,2,4,6,8,8)\), and blocks per level \((1,1,1,2,2,3,3,4)\). Local attention with window size \(128\) is used at the first three levels and global attention at coarser levels; shifted windows, RoPE, Fourier positional features, and SwiGLU are enabled. The noisy input is centered at \(0.5\) and scaled by \((\sigma^2+\sigma_{\mathrm{data}}^2)^{-1/2}\).

All binarized CIFAR-10 runs use batch size \(64\), \(1000\) epochs, EMA decay \(0.9997\), AdamW, learning rate \(2\times10^{-4}\), \((\beta_1,\beta_2)=(0.9,0.99)\), weight decay \(10^{-2}\), gradient clipping at \(1.0\), cosine learning-rate decay with warm-up, mixed precision where supported, PyTorch compilation, and FlashAttention-compatible kernels when available. Sampling uses \(128\) Heun steps, requiring \(255\) NFE, with \(\sigma_{\max}^{\mathrm{eval}}=20.0\) and \(\sigma_{\mathrm{decode}}=0.1\). We report FID on \(50{,}000\) generated samples.

\paragraph{OpenWebText-2 bitstreams.}
\label{app:owt_details}

For OpenWebText-2, we train a byte-level BPE tokenizer with vocabulary size \(32{,}768\). Tokens are ordered by a semantic traversal of skip-gram Word2Vec embeddings and encoded with Gray code. The resulting \(15\)-bit token code is expanded to \(21\) bits using SECDED-style error correction. Each example contains \(256\) tokens, giving \(D=5376\). We build a \(10^9\)-token cache and split it deterministically into train, validation, and test partitions with fractions \(0.99/0.005/0.005\). 
Training uses conditional prefix completion. Prefix length is sampled uniformly from \(0\) to \(128\) tokens. Prefix bits remain clean, suffix bits are corrupted, and the loss is evaluated only on the suffix. Classifier-free guidance is trained by dropping the prefix with probability \(0.2\) and replacing it with the null value \(0.5\). The model is a flat transformer over \(21\)-bit patches with content embedding dimension \(64\), trunk dimension \(512\), \(16\) transformer blocks, \(8\) attention heads, RoPE, AdaLN-Zero conditioning, local absolute positional side features, SwiGLU feed-forward layers, and self-conditioning probability \(0.5\). A skip-MLP head combines patch-level context with fine per-bit features and outputs Bernoulli logits for every bit. All OpenWebText-2 runs use batch size \(256\), \(66\) epochs, approximately \(10^6\) optimization steps, and EMA decay \(0.9999\). Optimizer and numerical settings match the bitstream CIFAR-10 runs. Sampling uses \(64\) Heun steps, requiring \(127\) NFE, and stops at \(\sigma=0.185\), where the expected bit MSE is approximately \(10^{-12}\). At inference, conditional and unconditional probability denoisers are combined before conversion to the VE score. We evaluate guidance scales \(\{0.0,1.5,2.0\}\) and report \(2.0\) in the main text. Generated bitstreams are thresholded, decoded by reversing ECC and semantic Gray coding, and converted to text with the trained tokenizer. We report MAUVE on \(10{,}240\) generated test samples using \texttt{gpt2-large} features, maximum text length \(256\), and micro-batch size \(512\).

\subsection{Metrics, inference budgets, and compute}
\label{app:metrics_compute}

Metrics are computed with the same inference sampler and NFE within each dataset. Image and bitstream-image experiments report FID from \(50{,}000\) generated samples. DNA experiments report Sei-FID. Text experiments report MAUVE on \(10{,}240\) generated samples from the test split using \texttt{gpt2-large} features. The NFE reported in the main table is the number of denoiser evaluations used by the evaluation sampler.

All reported experiments were run on NVIDIA H100 GPUs. CIFAR-10 and FFHQ \(64{\times}64\) used 8 GPUs per run; all other experiments used 4 GPUs on the same cluster. Main-text compute budgets are reported in processed examples. Wall-clock ranges are hardware-dependent references: CIFAR-10 took \(\sim\)2d 00h--2d 02h to 200kimg, FFHQ \(64{\times}64\) took \(\sim\)2d 09h--2d 13h to 200kimg, binarized CIFAR-10 took \(\sim\)7h--9h to 200kimg.

 \subsection{Existing assets and licenses}
\label{app:assets_licenses}

We use public datasets, model components, and evaluation tools. The original sources are cited below, and license or terms information is reported from the corresponding source when available. We do not redistribute any dataset as a new asset.

\begin{table}[h]
    \centering
    \caption{
    \textbf{Existing assets used in the experiments.}
    License and terms entries follow the corresponding source documentation where available.
    }
    \label{tab:assets_licenses}
    \begin{tabular}{p{0.25\linewidth}p{0.25\linewidth}p{0.42\linewidth}}
        \toprule
        Asset & Use in this paper & License / terms \\
        \midrule
        CIFAR-10~\citep{krizhevsky2009learning}
        & Image and bitstream experiments
        & Public benchmark dataset; original Toronto page describes the dataset but does not state an explicit license \\
        
        FFHQ~\citep{karras2019style}
        & FFHQ \(64{\times}64\) image experiments
        & Images retain their original Flickr licenses; dataset metadata and tooling are released under the terms stated by the FFHQ source \\
        
        OpenWebText-2~\citep{gao2020pile}
        & Text bitstream experiments
        & MIT license, as stated by the EleutherAI OpenWebText2 repository \\
        
        EDM code and setup~\citep{karras2022elucidating}
        & Image baseline, preconditioning, loss weighting, and sampler
        & Official NVIDIA research code release; repository includes a license file and directs licensing inquiries to NVIDIA Research Licensing \\
        
        MAUVE~\citep{pillutla2021mauve}
        & Text evaluation metric
        & GPLv3, as stated by the \texttt{mauve-text} package metadata \\
        
        GPT-2-large~\citep{radford2019language}
        & Feature extractor for MAUVE
        & Modified MIT License, as stated by the model release \\
        \bottomrule
    \end{tabular}
\end{table} 
\section{Allocation diagnostics and qualitative results}
\label{app:additional_diagnostics}

This appendix collects diagnostics that connect the online estimator to the allocation behavior used in the main experiments. We first show how unweighted denoising losses are converted into an information-profile estimate and then into an effective allocation. We then provide dataset-level profile galleries, qualitative samples at matched checkpoints, and a consistency argument for the binned loss estimate used during sampler rebuilds.

\begin{figure}[t]
    \centering
    \includegraphics[width=\linewidth]{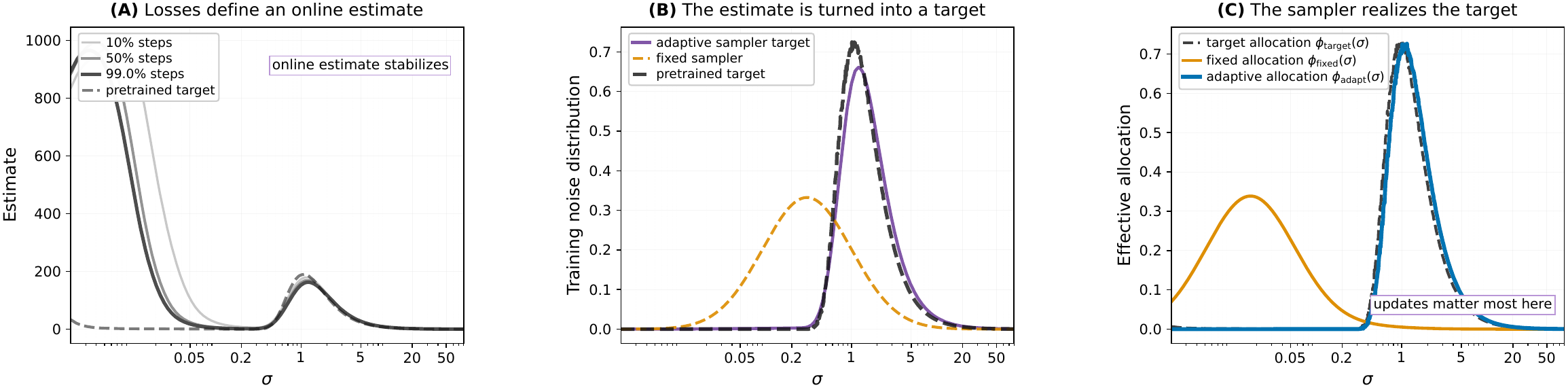}
\caption{
\textbf{\method{} turns online loss statistics into effective allocation.}
\textbf{(A)} Unweighted denoising losses measure the model's \(x\)-space error across noise levels.
\textbf{(B)} The path factor converts these losses into an online information-profile estimate, shown alongside a fixed reference sampler and a post hoc diagnostic profile.
\textbf{(C)} Rebuilding the sampler changes the induced effective allocation while the objective and loss weighting remain fixed.
}
    \label{fig:method_pipeline}
\end{figure}

\begin{figure*}[t]
  \centering
  \begin{subfigure}[t]{0.49\textwidth}
    \centering
    \includegraphics[width=\linewidth]{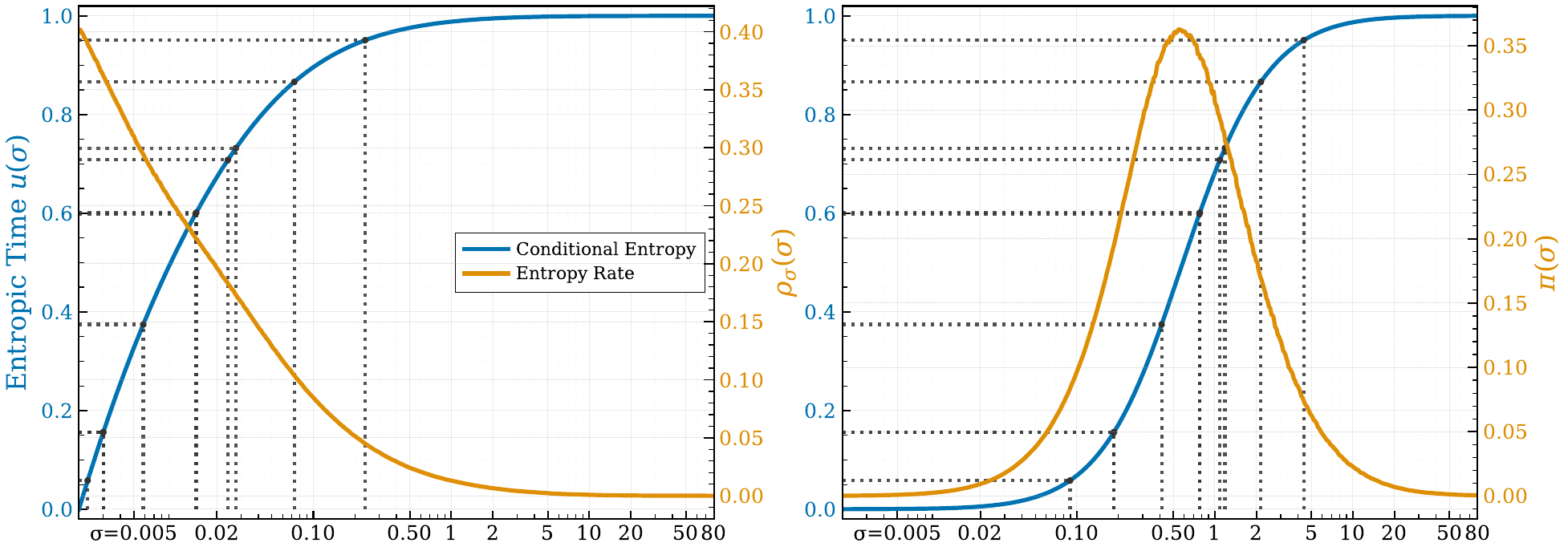}
    \subcaption{CIFAR-10}
  \end{subfigure}\hfill
  \begin{subfigure}[t]{0.49\textwidth}
    \centering
    \includegraphics[width=\linewidth]{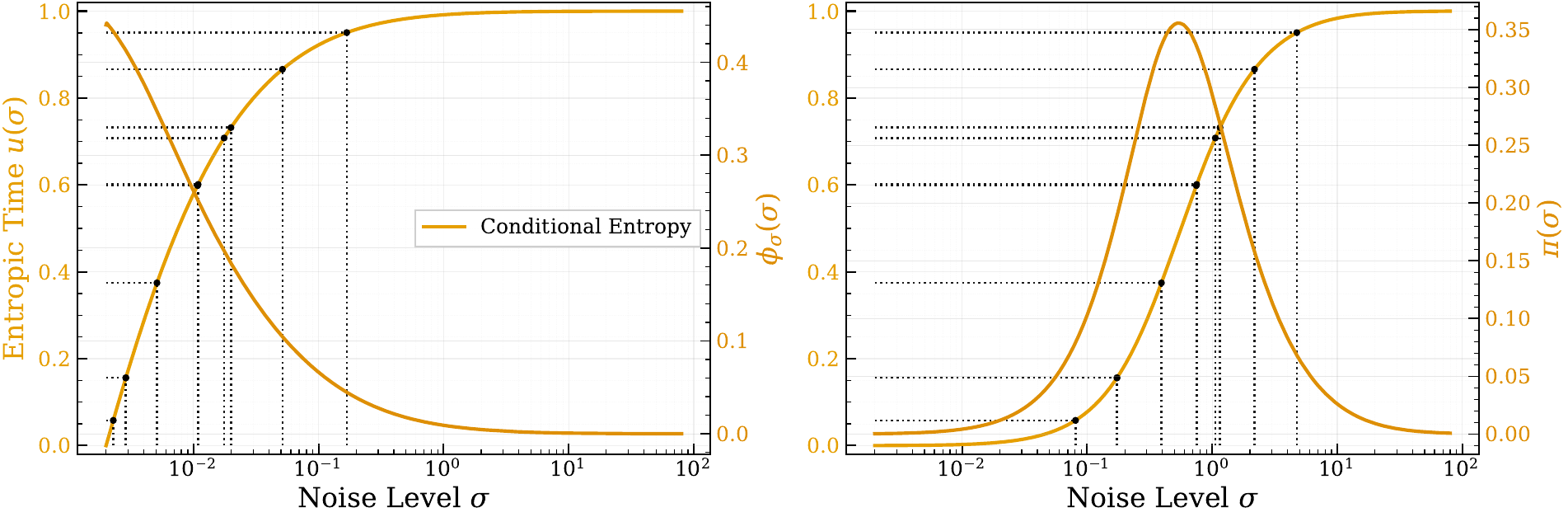}
    \subcaption{ImageNet-64}
  \end{subfigure}

  \vspace{0.35em}

  \begin{subfigure}[t]{0.49\textwidth}
    \centering
    \includegraphics[width=\linewidth]{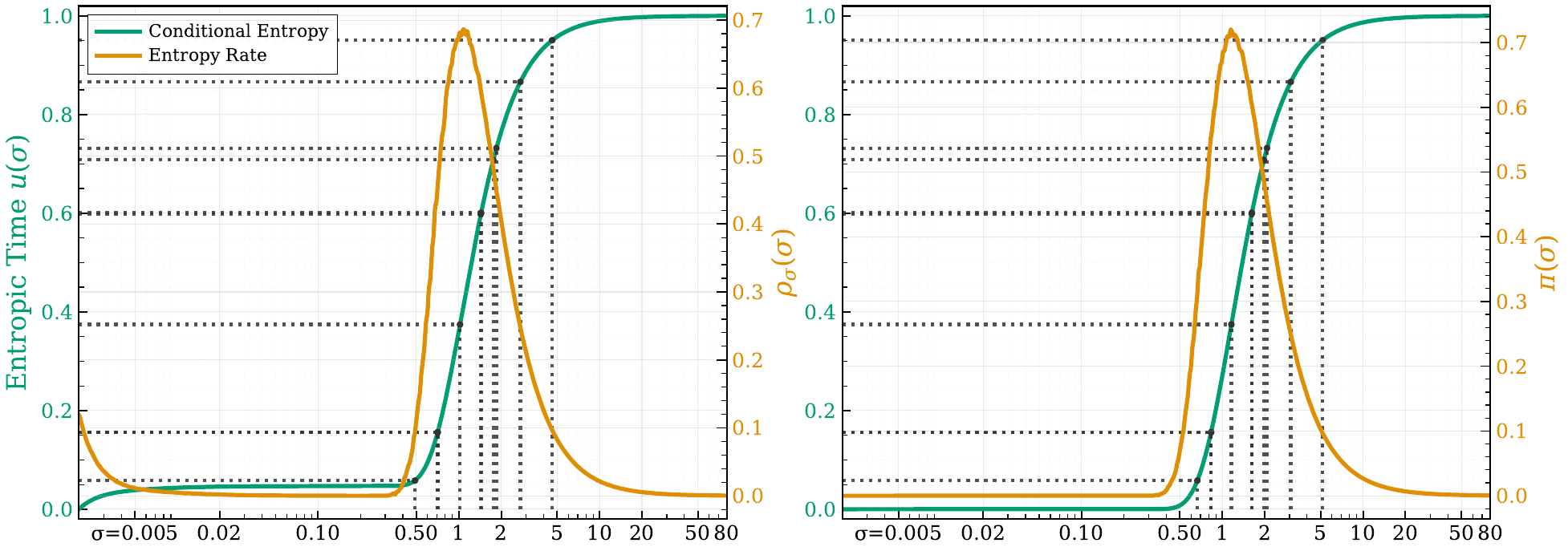}
    \subcaption{FFHQ-64}
  \end{subfigure}\hfill
  \begin{subfigure}[t]{0.49\textwidth}
    \centering
    \includegraphics[width=\linewidth]{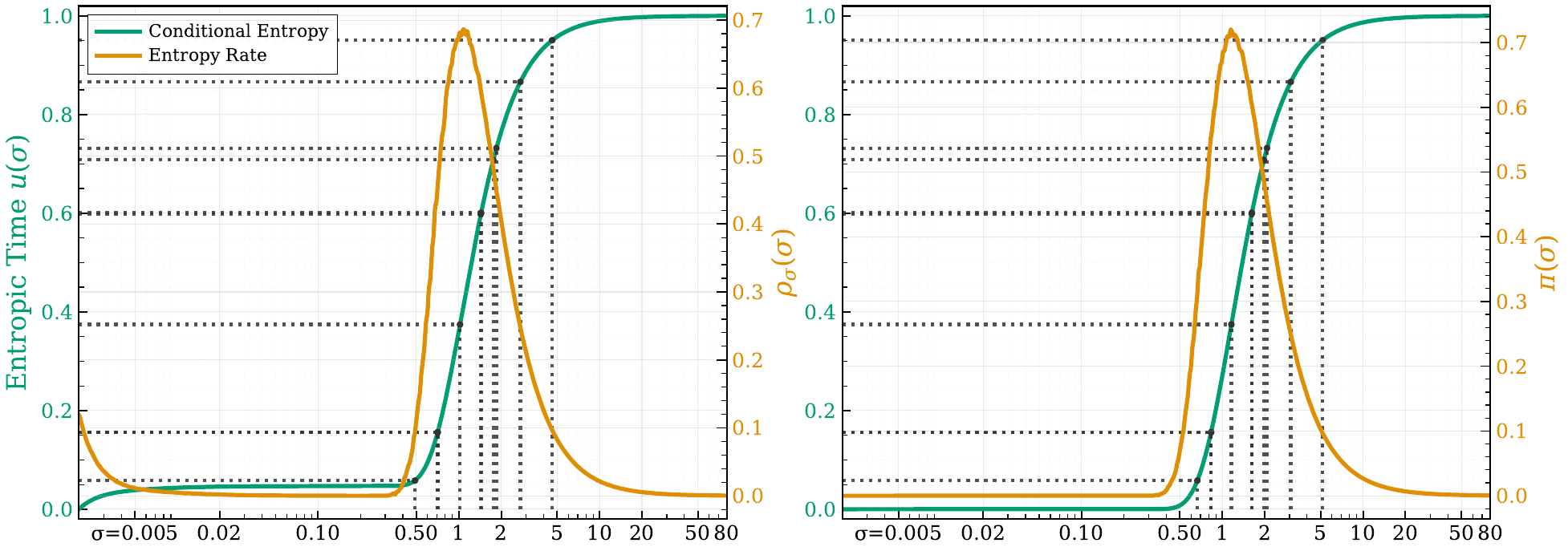}
    \subcaption{DNA}
  \end{subfigure}

  \vspace{0.35em}

  \begin{subfigure}[t]{0.49\textwidth}
    \centering
    \includegraphics[width=\linewidth]{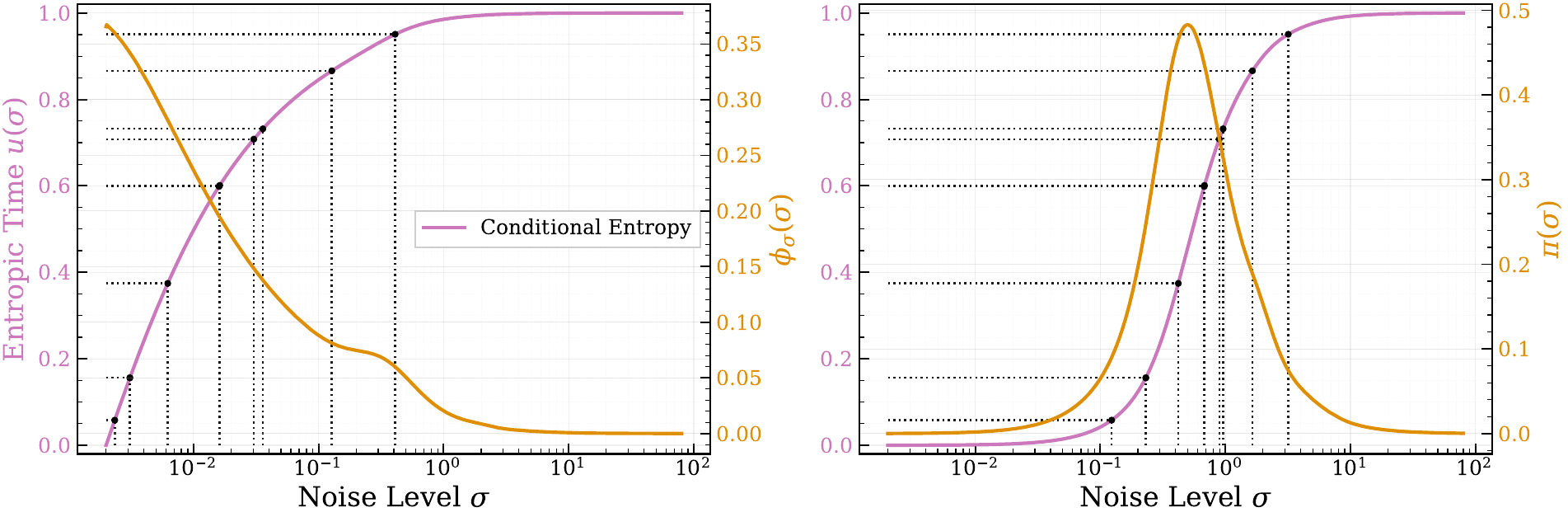}
    \subcaption{Binarized MNIST}
  \end{subfigure}\hfill
  \begin{subfigure}[t]{0.49\textwidth}
    \centering
    \includegraphics[width=\linewidth]{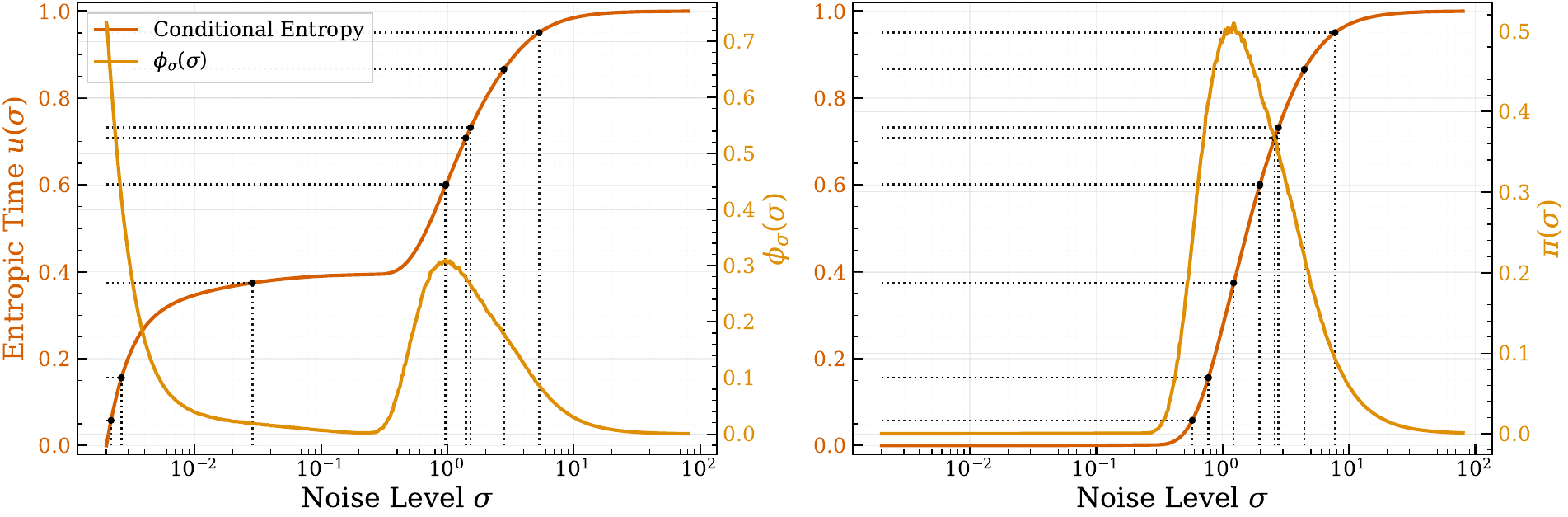}
    \subcaption{MNIST}
  \end{subfigure}

\caption{
\textbf{\method{} estimates dataset-specific information profiles.}
Each panel shows the normalized online profile estimate and the inverse-CDF sampler induced by \method{}. The sampler \(\pi(\sigma)\), together with the fixed per-noise weight \(w(\sigma)\), determines the realized allocation \(\phi(\sigma)=\pi(\sigma)w(\sigma)\). The profiles show that the informative region shifts across image, discrete, and sequence regimes.
}
  \label{fig:entropic_scheduler_gallery}
\end{figure*}

\begin{figure*}[t]
  \centering
  \begin{subfigure}[t]{0.49\textwidth}
    \centering
    \includegraphics[width=\linewidth]{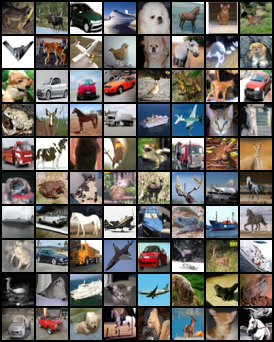}
    \subcaption{\textbf{CIFAR-10 @100k} \method{}, Heun, NFE\(=35\)}
  \end{subfigure}\hfill
  \begin{subfigure}[t]{0.49\textwidth}
    \centering
    \includegraphics[width=\linewidth]{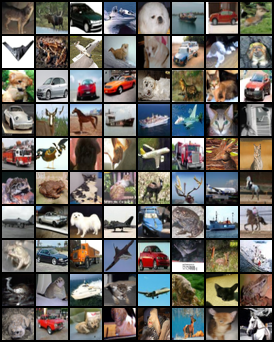}
    \subcaption{\textbf{CIFAR-10 @100k} EDM, Heun, NFE\(=35\)}
  \end{subfigure}

  \vspace{0.35em}

  \begin{subfigure}[t]{0.49\textwidth}
    \centering
    \includegraphics[width=\linewidth]{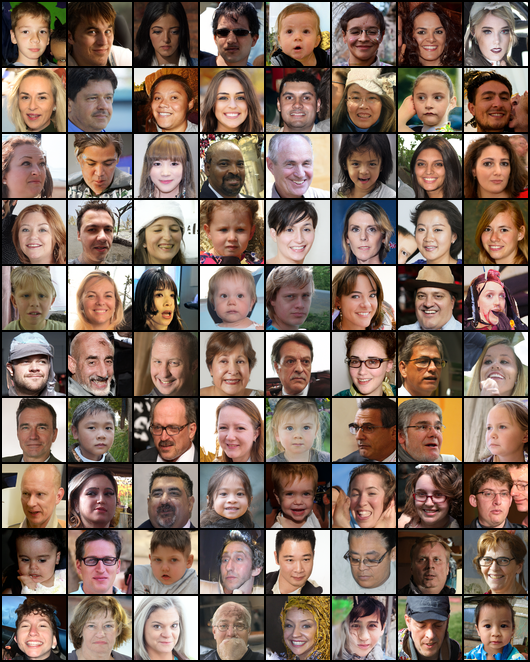}
    \subcaption{\textbf{FFHQ-64 @160k} \method{}, Heun, NFE\(=79\)}
  \end{subfigure}\hfill
  \begin{subfigure}[t]{0.49\textwidth}
    \centering
    \includegraphics[width=\linewidth]{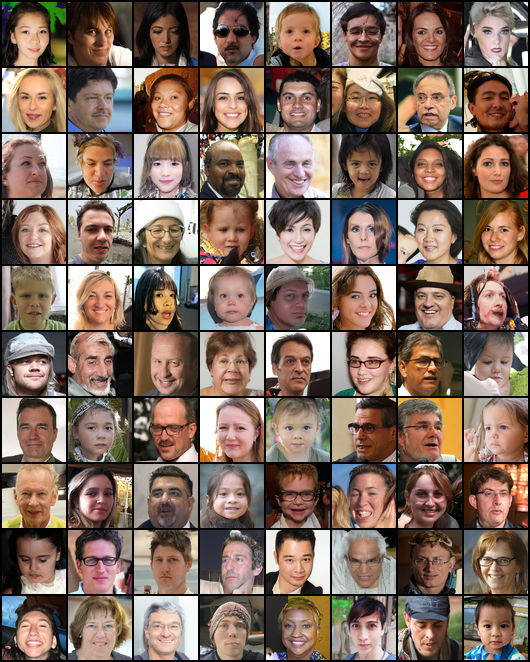}
    \subcaption{\textbf{FFHQ-64 @160k} EDM, Heun, NFE\(=79\)}
  \end{subfigure}
\caption{
\textbf{Qualitative samples at matched training checkpoints.}
\method{} and EDM are evaluated with the same solver, inference grid, and NFE. Checkpoints are matched by processed examples.
}
  \label{fig:app_qualitative_samples}
\end{figure*}

\subsection{Consistency of the binned loss estimate}
\label{app:binned_loss_consistency}

Fix a denoiser \(\hat x_\theta\), a sampler \(\pi\), and a bin \(B_j\subset\mathcal U\) with nonzero sampler mass. Define
\[
L_\theta(u)
=
\E\!\left[
\|x_0-\hat x_\theta(x_u;u)\|_2^2
\mid u
\right].
\]
If the conditional loss variance is finite, the empirical average of losses whose sampled noise levels fall in \(B_j\) converges almost surely to
\[
\E_{\pi}\!\left[L_\theta(u)\mid u\in B_j\right].
\]
As the grid is refined and \(L_\theta(u)\) is continuous, this conditional average converges to \(L_\theta(u_j)\) for any representative point \(u_j\in B_j\). If the denoiser is Bayes-optimal,
\[
\hat x_\theta(x_u;u)=\E[x_0\mid x_u],
\]
then \(L_\theta(u)=\mmse(\gamma(u))\). Multiplying by \(\frac12|\gamma'(u)|\) gives the conditional-entropy-rate profile up to normalization. In the online algorithm, this binned statistic is used as a plug-in estimate at each sampler refresh; smoothing, minimum bin counts, warm-up, and endpoint gating keep the finite-sample estimate stable while the model and sampler evolve.

Together, these diagnostics support the main claim that \method{} adapts the training distribution through an online estimate of where denoising uncertainty is resolved, rather than through post hoc profile access or a changed objective.



\end{document}